\PassOptionsToPackage{unicode=true}{hyperref} % options for packages loaded elsewhere
\PassOptionsToPackage{hyphens}{url}
\documentclass[12pt,]{article}
\usepackage{lmodern}
\usepackage{amssymb,amsmath}
\usepackage{ifxetex,ifluatex}
\usepackage{fixltx2e} % provides \textsubscript
\ifnum 0\ifxetex 1\fi\ifluatex 1\fi=0 % if pdftex
  \usepackage[T1]{fontenc}
  \usepackage[utf8]{inputenc}
  \usepackage{textcomp} % provides euro and other symbols
\else % if luatex or xelatex
  \usepackage{unicode-math}
  \defaultfontfeatures{Ligatures=TeX,Scale=MatchLowercase}
\fi
\IfFileExists{upquote.sty}{\usepackage{upquote}}{}
\IfFileExists{microtype.sty}{%
\usepackage[]{microtype}
\UseMicrotypeSet[protrusion]{basicmath} % disable protrusion for tt fonts
}{}
\IfFileExists{parskip.sty}{%
\usepackage{parskip}
}{% else
\setlength{\parindent}{0pt}
\setlength{\parskip}{6pt plus 2pt minus 1pt}
}
\usepackage{hyperref}
\hypersetup{
            pdftitle={Scale-invariant temporal history (SITH): optimal slicing of the past in an uncertain world},
            pdfauthor={Tyler A. Spears \^{}\{1\}; Brandon G. Jacques \^{}\{1\}; Marc W. Howard \^{}2; Per B. Sederberg \^{}\{1*\}},
            pdfborder={0 0 0},
            breaklinks=true}
\usepackage{graphicx,grffile}
\makeatletter
\def\maxwidth{\ifdim\Gin@nat@width>\linewidth\linewidth\else\Gin@nat@width\fi}
\def\maxheight{\ifdim\Gin@nat@height>\textheight\textheight\else\Gin@nat@height\fi}
\makeatother
\setkeys{Gin}{width=\maxwidth,height=\maxheight,keepaspectratio}
\ifx\paragraph\undefined\else
\let\oldparagraph\paragraph
\renewcommand{\paragraph}[1]{\oldparagraph{#1}\mbox{}}
\fi
\ifx\subparagraph\undefined\else
\let\oldsubparagraph\subparagraph
\renewcommand{\subparagraph}[1]{\oldsubparagraph{#1}\mbox{}}
\fi

\makeatletter
\def\fps@figure{htbp}
\makeatother

\usepackage[utf8]{inputenc}
\usepackage[T1]{fontenc}
\usepackage{hyperref}
\usepackage{url}
\usepackage{booktabs}
\usepackage{amsfonts}
\usepackage{nicefrac}
\usepackage{microtype}
\usepackage[top=1in, bottom=1in, left=1.25in, right=1.25in]{geometry}

\title{Scale-invariant temporal history (SITH): optimal slicing of the past in
an uncertain world}
\author{Tyler A. Spears \(^{1}\) \and Brandon G. Jacques \(^{1}\) \and Marc W. Howard \(^2\) \and Per B. Sederberg \(^{1*}\)}
\date{}

\begin{document}
\maketitle
\begin{abstract}
In both the human brain and any general artificial intelligence (AI), a
representation of the past is necessary to predict the future. However,
perfect storage of all experiences is not feasible. One approach
utilized in many applications, including reward prediction in
reinforcement learning, is to retain recently active features of
experience in a buffer. Despite its prior successes, we show that the
fixed length buffer renders Deep Q-learning Networks (DQNs) fragile to
changes in the scale over which information can be learned. To enable
learning when the relevant temporal scales in the environment are not
known \emph{a priori}, recent advances in psychology and neuroscience
suggest that the brain maintains a compressed representation of the
past. Here we introduce a neurally-plausible, scale-free memory
representation we call Scale-Invariant Temporal History (SITH) for use
with artificial agents. This representation covers an exponentially
large period of time by sacrificing temporal accuracy for events further
in the past. We demonstrate the utility of this representation by
comparing the performance of agents given SITH, buffer, and exponential
decay representations in learning to play video games at different
levels of complexity. In these environments, SITH exhibits better
learning performance by storing information for longer timescales than a
fixed-size buffer, and representing this information more clearly than a
set of exponentially decayed features. Finally, we discuss how the
application of SITH, along with other human-inspired models of
cognition, could improve reinforcement and machine learning algorithms
in general.
\end{abstract}

\textbf{Affiliations}

\(^1\) Department of Psychology, University of Virginia, USA\\
\(^2\) Department of Psychology, Boston University, USA\\
\(^*\) Corresponding Author, Per B. Sederberg, 434-924-5725 (phone),
pbs5u@virginia.edu (email)

\hypertarget{introduction}{%
\section{Introduction}\label{introduction}}

In many situations, a natural or artificial learner is faced with a
continuous stream of experience. It is often advantageous to learn
dependencies between causes and outcomes that are (perhaps widely)
separated in time from one another. In reinforcement learning (RL)
applications, an agent must learn to take actions in the present to
optimize rewards in the future (Mnih et al., 2015). A representation of
the past is necessary if one wishes to effectively predict the future
(Ba, Hinton, Mnih, Leibo, \& Ionescu, 2016). In this paper, we consider
three models for representing the past: a fixed-size buffer, a set of
exponentially-decayed features, and our proposed Scale-Invariant
Temporal History (SITH) model. To analyze their capabilities, we compare
the performance of artificial agents given each model in two example
video game environments.

Choosing a representation of history is nontrivial, and there are at
least two difficulties in selecting such a memory mechanism. The first
difficulty is balancing the cost of memory storage with accuracy of
prediction. To date, there have been two main approaches to history
representation: maintaining a fixed buffer size of information into the
past (typically implemented as a first-in first-out {[}FIFO{]} buffer of
length \(N\)), or maintaining graded feature activations that decay
exponentially. These options are not unlike the long-standing debate
arising from laboratory-based studies in the field of human memory as to
the nature of short-term or working memory representations (P. B.
Sederberg, Howard, \& Kahana, 2008).

The second difficulty is ensuring flexibility and generality in the
representation. Unlike most laboratory-based studies, the real world
manifests temporal relationships on an incredible diversity of time
scales. Causes predict outcomes on the scale of milliseconds, seconds,
hours, days, and so on. However, an RL model constructed from a FIFO
buffer of length \(N\), or even an exponentially-decaying graded
representation, introduces a characteristic scale. By introducing a
particular scale, one reduces the general capabilities of the model by
forcing it to operate under that scale. Specifically, the model will
have qualitatively different behavior when the temporal relationships to
be estimated in the world are less than \(N\) compared to when the
relationships are greater than \(N\) (Z. Tiganj, Gershman, Sederberg, \&
Howard, in press). One might address this concern by choosing \(N\) to
be large, but this choice has major implications for the amount of
resources the agent consumes during its decision-making process. The
number of buffer nodes to represent a particular scale \(\tau\) goes up
linearly with \(\tau\); in neural network applications, the number of
weights can go up as rapidly as \(\tau^2\). Successfully choosing \(N\)
in a resource-conserving way requires us to know \emph{a priori} the
temporal relationships we are likely to observe. In practice, this
requires us to make critical assumptions about the problem we are trying
to learn \emph{prior} to training the model.

\begin{figure}
\centering
\includegraphics[width=0.73\textwidth,height=\textheight]{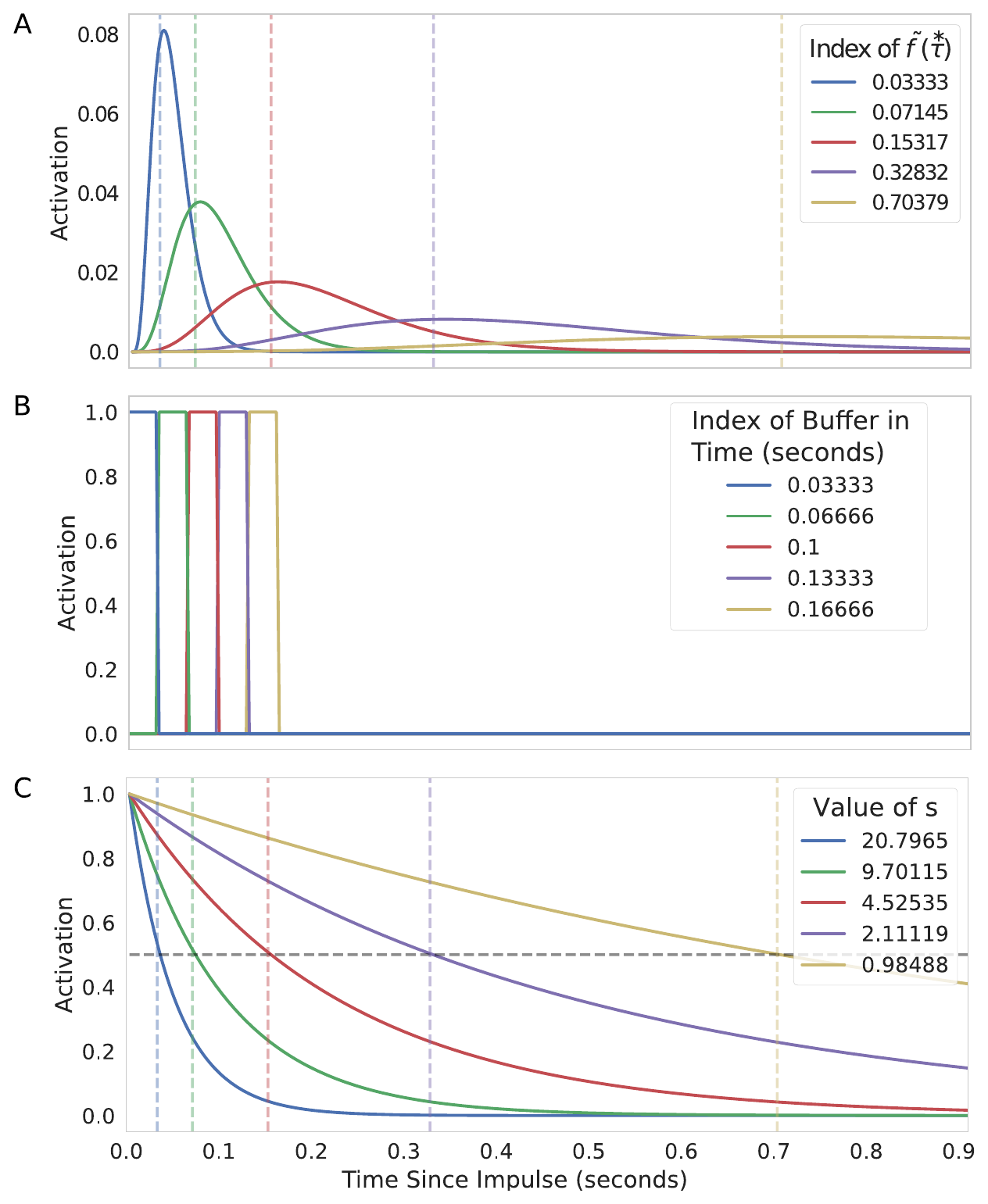}
\caption{\emph{SITH efficiently extends into the past, unlike a buffer,
and retains temporal order information more clearly than exponential
decay.} For all visuals presented here, a single impulse was presented
for 0.0033 seconds, followed by a decay of 0.9 seconds; the activation
values of each index for each representation are shown through time. (A)
Activation of SITH through time. By choosing five evenly spaced indices
from the full output of SITH, we retrieve a sparse, compressed
representation of a single feature following its transient activation;
each line represents an index in this subset, which is a temporal
receptive field centered on the corresponding dashed, vertical line.
Note the logarithmic spacing of centers, as well as the asymmetric,
increased spread of activation of fields centered further in time. (B)
Activation of a FIFO buffer through time. A buffer of size 5 has
perfectly accurate storage of the impulse, but does not span long
distances in time. Here, each buffer index stores one frame of game
input, equivalent to \(1/30^{\text th}\) of a second. (C) Activation of
exponentially-decaying representations through time. Each line denotes
the activation of a different decay rate through time. The decay rates
\(s\) were chosen to have an activation at 0.5 at the matching temporal
receptive fields in SITH. While the exponential decay does span a large
amount of time, there is little information indicating \emph{when} the
impulse occurred, just that it happened at some
point.\label{fig:rep_activation}}
\end{figure}

Our goal is to provide a memory representation that tackles both
difficulties. Below, we describe and characterize this model, called the
Scale-Invariant Temporal History (SITH). Then, we consider the
implications of utilizing SITH in place of either a FIFO buffer of size
\(N\) or a family of exponentially decaying representations, all of
which are illustrated in Figure \ref{fig:rep_activation}. The
capabilities of each representation are compared through manipulations
of two reward-driven, arcade-style games. Finally, we compare the
performance of all three representations, and discuss how SITH can
enrich memory systems in AI.

\hypertarget{scale-invariant-temporal-history-an-efficient-representation-for-learning-across-scales}{%
\section{Scale-Invariant Temporal History: An efficient representation
for learning across
scales}\label{scale-invariant-temporal-history-an-efficient-representation-for-learning-across-scales}}

SITH builds on a body of work in neural computation and cognitive
psychology, as found in K. H. Shankar \& Howard (2013) and Howard,
Shankar, Aue, \& Criss (2015), along with neurological evidence for its
existence in mammalian brains detailed in M. W. Howard et al. (2014) and
Z. Tiganj, Cromer, Roy, Miller, \& Howard (2018). The method results in
a logarithmically compressed representation that coarse-grains the
history of what happened, and when. With this logarithmic compression,
the number of nodes necessary to represent a time \(\tau\) in the past
goes up only like \(\log\tau\): an exponential savings. The cost is that
the resolution at which past events can be discriminated goes down
linearly with the distance into the past. This decrement in accuracy
obeys the Weber-Fechner law, a fundamental result in human and animal
psychophysics (Fechner, Howes, \& Boring, 1966). Furthermore, this form
of coarse-graining can be shown to be an optimal solution to processing
signals with a particular scale, but where the agent uses an
uninformative prior for the unknown scale (Howard \& Shankar, 2018).

\hypertarget{model-definition}{%
\subsection{Model Definition}\label{model-definition}}

For each feature in an agent's observation, one can imagine a set of
units that have temporal receptive fields. The units that fire when a
feature was active closest to the present have smaller receptive fields
that are centered closer together, much like the fovea of the retina
(Van Essen, Newsome, \& Maunsell, 1984). These units provide high
temporal accuracy, but are activated for only a small time into the
past. Conversely, neurons that are active further away from the present
have larger receptive fields that are spaced further apart. In keeping
with prior work (K. H. Shankar \& Howard, 2013), the nodes of this
representation are evenly spaced on a logarithmic scale.

To attain this SITH representation from a stream of input, we first
compute a set of leaky integrators
\[\frac{dF(s)}{dt } = -s F(s) + f(t),\] where each unit is indexed by
its value of \(s\) and driven by the function \(f(t)\) that describes
the activation of a particular feature (in our case pixels on a screen).
It can be shown that this set of leaky integrators implements the
Laplace transform of the input function over past times \(f(\tau < t)\).
We approximate the inversion of the Laplace transform using a linear
operator \(\mathbf{L}^{-1}_k\) that provides a discrete approximation to
the Post inversion formula (K. H. Shankar \& Howard, 2013).

The goal of this method is to approximate the history of the input
function leading up to the present \(f(\tau < t)\). We make this
estimate by approximating the inversion of the Laplace transform of this
history. For each unit in \(F(s)\), we assign a new unit in
\(\tilde{f}(\overset{*}{\tau})\), where \(s\) and \(\overset{*}{\tau}\)
are in one-to-one correspondence as \(s = -k/\overset{*}{\tau}\). At
each moment we compute
\[\tilde{f}(\overset{*}{\tau}) = \mathbf{L}^{-1}_k F(s).\] The integer
\(k\) controls the degree of approximation. The error in the inversion
introduces a coarse-graining function that depends on the value of \(k\)
(for these simulations \(k=4\).) As \(k \rightarrow \infty\) the
inversion becomes perfect. If we refer to the present as \(t=0\), a node
with a particular value of \(\overset{*}{\tau}\) provides a
coarse-grained estimate of the true function, \(\overset{*}{\tau}\) in
the past: \(\tilde{f}(\overset{*}{\tau}) \simeq f(\overset{*}{\tau})\),
with equality as \(k \rightarrow \infty\). However, the coarse-graining
is scale-invariant; the error in the reconstruction can be shown to be
proportional to \(\overset{*}{\tau}\). By choosing \(s\) such that the
\(\overset{*}{\tau}\) values are logarithmically-compressed, we
construct a scale-invariant temporal history. One approach to
parameterizing \(s\) (and, consequently, \(\overset{*}{\tau}\)) is

\[s_i = -k/\overset{*}{\tau}_i = \overset{*}{\tau}_0 (1 + c)^{i} ,\]

where \(0 \leq i \leq N_{\overset{*}{\tau}} - 1\), and
\(N_{\overset{*}{\tau}} \geq 2k +1\). This definition is used in both
these simulations, and previous work (K. H. Shankar \& Howard, 2013).
Due to the one-to-one correspondence between \(s\) and
\(\overset{*}{\tau}\), note that
\(\lvert s \rvert = \lvert \overset{*}{\tau} \rvert = N_{\overset{*}{\tau}}\).
Note that the representation at a particular value of
\(\overset{*}{\tau}\) is independent of other values of
\(\overset{*}{\tau}\).

The evolution of the representation depends only on the values of
\(F(s)\) in the neighborhood of \(-k/\overset{*}{\tau}\). This means
that although the representation can be thought of as continuous, we can
sample whatever discrete values we find convenient for a particular
application. In the applications described here, we take only four or
five (dependent on the environment) discrete values and use it in place
of traditional FIFO buffers and sets of exponentially decaying features.
A comparison of these representations is illustrated in Figure
\ref{fig:rep_activation}.

\hypertarget{parameter-characterization}{%
\subsection{Parameter
Characterization}\label{parameter-characterization}}

SITH is governed by relatively few parameters, though each can have
subtle effects on the resulting representation. As discussed above,
\(N_{\overset{*}{\tau}}\) sets the size of the ensemble in \(F(s)\).
Additionally, there are three other parameters that determine the
temporal scale, spacing, and fidelity of each value in
\(\tilde{f}(\overset{*}{\tau})\), which are \(\overset{*}{\tau}_0\),
\(c\), and \(k\).

\begin{figure}
\centering
\includegraphics{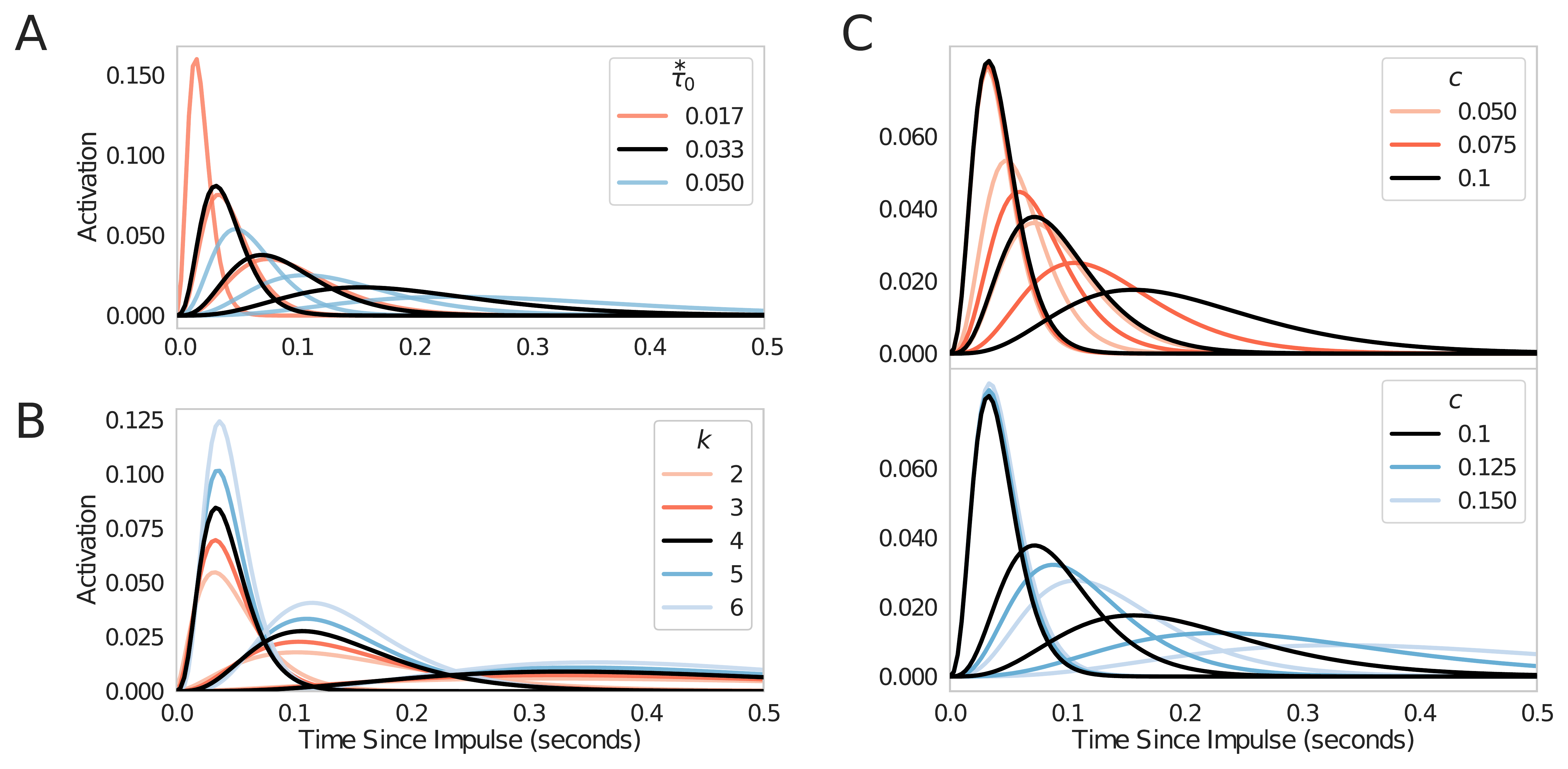}
\caption{\emph{Three parameters control SITH's minimum scale, fidelity,
and spacing.} For all figures, an impulse with activation 1 was
presented for 0.033 seconds, followed by a delay of 0.5 seconds. Unless
otherwise indicated, all parameters are equivalent to those used in the
first set of simulations; the black line in each plot corresponds to the
value used in these same simulations. After indexing linearly into
\(\tilde{f} ( \overset{*}{\tau} )\), only the first 3 values are shown
for clarity. (A) Modulation of \(\overset{*}{\tau}\) scaling. This
parameter sets the value of the first index in \(\overset{*}{\tau}\),
and constantly scales subsequent values. This can be seen in the
constant shift of activation peaks left and right, for smaller and
larger values, respectively, of \(\overset{*}{\tau}\). (B) Modulation of
degree of approximation \(k\). By increasing \(k\), more values from
\(F(s)\) are integrated in the \(\mathbf{L}^{-1}_k\) operator, creating
a more precise approximation of \(f(\overset{*}{\tau})\). This produces
stronger peaks in activation, at the cost of an increase in number of
operations performed and required size of \(s\). (C) Modulation of \(c\)
spacing. Parameter \(c\) determines the spacing of the \(s\) values. As
shown here, changes in \(c\) result in a logarithmic ``shift'' in the
peak activations. This nonlinear shift increases logarithmically, unlike
the constant shift from modifying \(\overset{*}{\tau}\). For clarity,
the plot is divided with descending values in the upper panel and
ascending values in the lower.\label{fig:params}}
\end{figure}

As given in our approach to constructing \(s\), \(\overset{*}{\tau}_0\)
acts as the value of the first \(\overset{*}{\tau}\) found in
\(\tilde{f}(\overset{*}{\tau})\). This can also be seen as the selection
of the timescale captured in the first index of
\(\tilde{f}(\overset{*}{\tau})\), with logarithmically increasing scales
in subsequent indices. This is visualized as a linear ``shift'' in the
peaks of each index in \(\tilde{f}(\overset{*}{\tau})\), with an
accompanying spreading of activation; this is shown in Figure
\ref{fig:params}A. For these simulations, we chose
\(\overset{*}{\tau}_0\) to be equal to one timestep in our discrete time
series, and we recommend this approach for all discrete time series
data.

The parameter \(c\), which controls the spacing of the \(s\) values,
also results in a ``shift'' of activation peaks, but a shift that is
logarithmically increasing. With all other parameters fixed, this causes
a relatively small change in the early \(\tilde{f}(\overset{*}{\tau})\)
indices, but can drastically change the spread of activation later in
the ensemble. Practically speaking, \(c\) can be tuned to maximize
SITH's temporal spread, while balancing the desired information
redundancy. The effect of modulating \(c\) is illustrated in Figure
\ref{fig:params}C. In these simulations, we set \(c=0.1\), which we
recommend for general usage.

Finally, the positive integer \(k\) determines the degree of
approximation in the Post approximation of the inverse Laplace
transform, \(\mathbf{L}^{-1}_k\). For any feature in some
\(\overset{*}{\tau}_{i}\), increasing \(k\) increases the number of
surrounding \(\overset{*}{\tau}\)'s that are used in the linear
combination to calculate \(\tilde{f}(\overset{*}{\tau}_i)\). Values of
\(k\) are directly proportional to SITH's accuracy in recreating past
events, at the cost of increasing the number of calculations performed
and the required number of \(s\) values. Increasing \(k\) causes the
peaks in each index of \(\tilde{f}(\overset{*}{\tau})\) to become
sharper, and more defined, as the approximation becomes more accurate;
this is illustrated in Figure \ref{fig:params}B. Here, as in other
works, we have \(k=4\), and we recommend this as a general default
value.

\hypertarget{testing-environments}{%
\section{Testing Environments}\label{testing-environments}}

To test the demands of maintaining information over a range of
timescales on a learning agent, we required a task that is solvable by
human agents, presents stimuli through time, and has a meaningful reward
structure. As demonstrated in Mnih et al. (2015), video games meet all
of these criteria.

\hypertarget{catch-a-simple-application-to-evaluate-the-ability-to-learn-and-express-temporal-relationships.}{%
\subsection{Catch: A simple application to evaluate the ability to learn
and express temporal
relationships.}\label{catch-a-simple-application-to-evaluate-the-ability-to-learn-and-express-temporal-relationships.}}

To reduce complexity and focus on the properties of the various
representations of interest, we first tested our agents in a noise-free,
simple video game environment that would not require convolution layers
to extract meaningful features. Catch, modified slightly from previously
published versions, provided a simple testbed to illustrate the
essential advantages of a scale-invariant representation (Ba et al.,
2016).

\begin{figure}
\centering
\includegraphics[width=1\textwidth,height=\textheight]{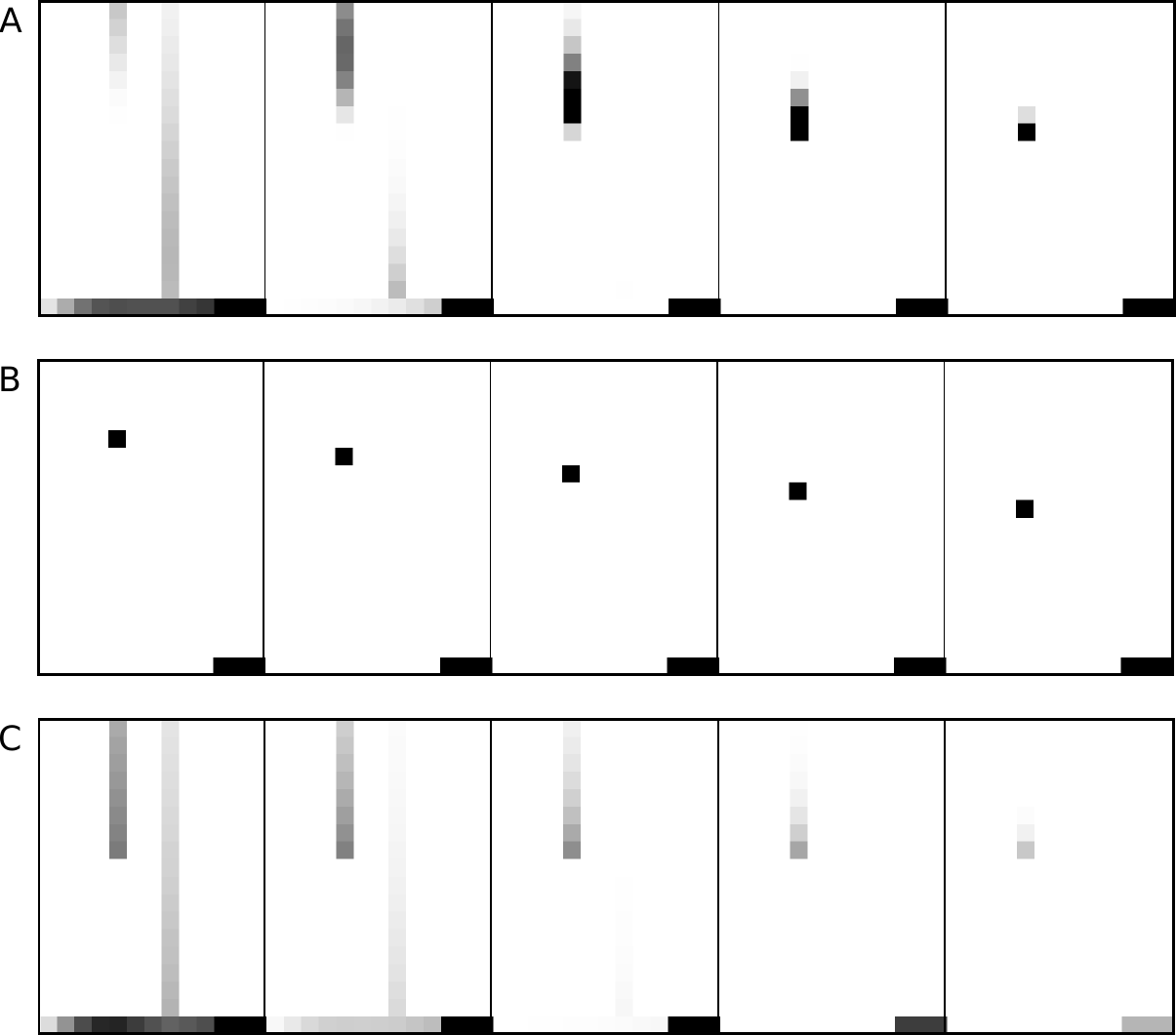}
\caption{\emph{SITH produces an extended, but focused, recreation of
events in pixel space compared to a buffer and exponential decay.}
Visualization of all representations tested with Catch, at the same
point in the same game. The most recent input is represented in the
rightmost index, and each representation spans further back in time
going leftwards. Colors are inverted and normalized within
representation, for clarity, such that darker values show stronger
activations. (A) Visualization of SITH given states of Catch. SITH spans
logarithmically increasing points into the past, while maintaining
information of when a feature was active. This can be seen by the shift
in intensity of the ball pixels, which follows the past, true path of
the ball. (B) Visualization of a buffer given states of Catch. While
information is perfectly preserved, that information can only span a
short discrete interval into the past depending on the length of the
buffer. (C) Visualization of exponential decay given states of Catch.
While the representation shown here spans further back than SITH, it is
unclear \emph{when} each feature was active in the past. This can be
seen by the most recent position of the ball having the highest
activation (in regards to pixels outside of the basket) in all indices.
This does not provide as useful a recreation of past experience, only
the knowledge \emph{that} a feature was active at some point in the past
relying on tiny relative changes in activation to estimate when it might
have occurred.\label{fig:rep_viz}}
\end{figure}

In Catch, an agent moves a basket left and right at the bottom of the
game screen in order to catch falling balls that start at the top of the
screen. Our instantiation of Catch is a game board comprising an
\(18 \times 13\) matrix of binary pixels; here, 1 represents the
presence of a game piece, and 0 represents the absence of any game
pieces. At the bottom of the game board is a \(1 \times 3\) basket.
Balls are spawned at the top of the game board, with only one ball on
the board at a time, and start to move down vertically to the bottom at
a constant rate of one pixel each frame. This is visualized most clearly
in Figure \ref{fig:rep_viz}B. The agent receives a 1 point reward if the
ball overlaps with the basket when the ball reaches the bottom row. If
the ball does not overlap, then the agent receives a -1 point
punishment. During every frame, the agent has to make a decision to
either move the basket one pixel to the left, one pixel to the right, or
to not move at all. For most of the tests, the game ends after 10 balls
fall to the bottom row, regardless of whether they were caught; as
explained below, we also performed limited testing with a variant of
Catch with only 1 ball per game, with otherwise identical task
parameters. New balls are spawned after the previous ball reaches the
bottom of the screen, so there is only a maximum of 1 ball on the game
screen at all times during training and testing. With the consideration
that the basket is 3 pixels wide, the screen is 13 pixels wide, and
there are 10 balls for each game, it is important to note that chance
performance for an agent performing random movements is approximately
\(-5\).

A simple modification to this game was made to demonstrate the effects
of obscuring parts of the stimuli, similar to the partially observable
version of Catch found in Ba et al. (2016). This variant, which we call
Hidden Catch, retained all properties of Catch as described above, with
one exception. In the game screen, a mask of varying size was placed
over all pixels in a set amount of rows, starting from the second to
bottom-most row. This mask set these pixels' values to 0, removing all
information for the agent. The agent was able to see and move the basket
in the same manner as before, however, the ball disappeared after
passing behind the mask. The agent received no further information about
the ball until the reward or penalty was given. In theory, so long as
the agent can store information about the horizontal position of the
ball, an optimal score can be attained. However, by increasing the size
of the mask, the agent must have the necessary memory representation to
learn this association and attain robust performance. This leads to the
fundamental computational challenge posed by Hidden Catch, namely that
the model must bridge larger and larger temporal gaps in order to behave
optimally.

\hypertarget{flappy-bird-a-more-complex-time-dependent-environment}{%
\subsection{Flappy Bird: A more complex time-dependent
environment}\label{flappy-bird-a-more-complex-time-dependent-environment}}

\begin{figure}
\centering
\includegraphics[width=0.8\textwidth,height=\textheight]{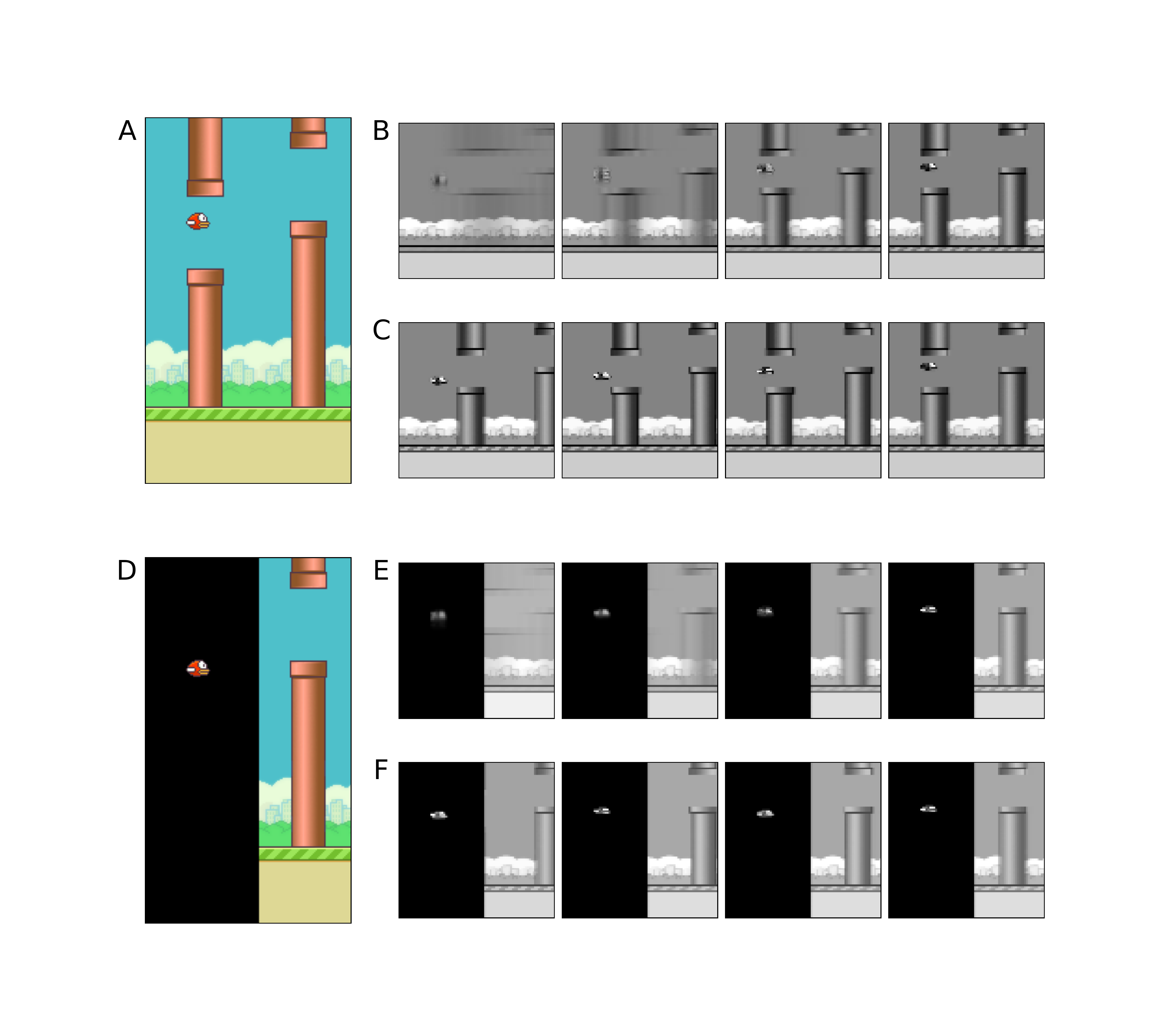}
\caption{\emph{SITH maintains rewarding information in a more complex
game.} (A) Illustration of a single screen of the visible variant of
Flappy Bird. The original screen size is \(288 \times 512\) pixels. (B)
Output of the SITH representation when given grayscaled and down-sampled
input. The most recent experience is maintained in the rightmost cell,
and spans further back into the past going left. (C) Output of a buffer
representation when given grayscaled and down-sampled input. The most
recent experience is in the rightmost cell, and experiences reach
further into the past going left. (D) Illustration of the partially
obscured variant of Flappy Bird, at the time of reward. The bird avatar
needs to remain visible, otherwise the task becomes impossible. (E)
Output of SITH representation given grayscaled and down-sampled input in
partially obscured Flappy Bird, at the time of reward. Note how the item
that leads to reward and punishment, the pipes, remains visible in the
cell spanning the most amount of time. (F) Output of the buffer
representation, given grayscaled and down-sampled input. Note how the
most important features pertaining to reward, the pipes nearest to the
avatar, are completely obscured at the time of
reward.\label{fig:flappy_bird}}
\end{figure}

After detailing the behavior of SITH in a simple, noise-free
environment, we sought to test in a higher-dimensional, noisier setting
to demonstrate that SITH can be combined with other standard
deep-learning approaches, namely convolutional layers. Our chosen game
was an emulated version of the game Flappy Bird (Tasfi, 2016). In Flappy
Bird, the goal of the agent is to navigate its character avatar between
pipes while constantly being pushed rightward. The agent can either
``flap'', leading to a gain in upward momentum, or perform a
``do-nothing'' operation, letting gravity pull the avatar down. Once the
avatar makes it through the gap in the pipes, the agent receives a
reward of 1. If the avatar touches the ground, the top of the screen, or
the pipes, the game is over, and the agent receives a penalty of -5. The
game is functionally endless, so an optimal agent would seek to go
between as many pipes as possible, for as long as possible. As seen in
Figure \ref{fig:flappy_bird}A, this environment has many more features
than Catch (7,056 features valued between 0 and 1, as opposed to 234
binary features, in these experiments). Critically, Flappy Bird also
contains irrelevant features that simply add noise to the reward
prediction (i.e.~background details, like bushes).

Similar to Hidden Catch, we also transformed Flappy Bird into a
partially-observable task with state correlations through time, by
masking pixels. This variant of Flappy Bird removes information of all
features behind and in front of the agent avatar. Critically, this mask
hides the location of the pipes at the time of reward or punishment. The
agent is only able to see the position of the avatar, and the normal
environment details beyond a certain point from the avatar. Thus, as
illustrated in Figure \ref{fig:flappy_bird}D, the agent must retain a
memory of where the pipes were, as well as how long ago they last saw
them, in order to flap at the correct time to dodge them.

\hypertarget{models-tested}{%
\section{Models Tested}\label{models-tested}}

\hypertarget{models-for-playing-catch}{%
\subsection{Models for Playing Catch}\label{models-for-playing-catch}}

As shown in Mnih et al. (2015), Deep Q-learning Networks (DQN) are
extraordinarily adept at solving simple video game environments. In
order to provide the DQN with some sense of temporal change, previous
applications have used a FIFO buffer of experience as input to the
network. The FIFO buffer keeps a perfectly accurate history of the past
\(N\) game frames. For Catch specifically, a visualization of a buffer
of size 5 during a moment of gameplay is shown in Figure
\ref{fig:rep_viz}B. The contents of the buffer are concatenated,
flattened, and passed into the network via the input layer. Thus, the
input to the network is of size \(N \times S\), where \(S\) is the size
of the stimulus set (game screen) at a single time point. During our
simulations in Catch, each game screen was \(13 \times 18\) pixels. We
tested the performance of a buffer of sizes 1, 5, and 10, resulting in
inputs of sizes 234, 1170, and 2340 features, respectively. We tested
these representations on a fully visible version of Catch, as well as
versions of Hidden Catch with logarithmically-spaced mask sizes of 1, 2,
4, 8, and 16.

Besides a buffer, another approach taken in the RL literature is the use
of exponentially-decaying sets of features, such as in Doya (2000).
Indeed, this approach is similar to the first process performed in SITH,
the calculation of the Laplace transform of the input. To capture some
given timespan, the decay rate is lowered between subsequent feature
sets, maintaining activations from stimuli further back in the past. We
found it informative to test the performance of an
exponentially-decaying representation, for two reasons: 1) as an analog
for the models described in previous work, and 2) as a justification for
the second process in the SITH model, the inverse Laplace transform from
exponentially-decaying representations into a logarithmically-compressed
representation of what features were active when in the past. Recalling
the parameters that compose SITH, we chose values of the decay rate
\(s\) that would reach half of their initial activation values at the
time of each \(\overset{*}{\tau}\) used in our instantiation of SITH;
the nature of these decaying features is shown in Figure
\ref{fig:rep_activation}C. This resulted in
\(s = [20.7965, 9.70115, 4.52535, 2.11119, 0.98488]\). The 5 decayed
feature representations are analogous to that of 5 stored experiences in
a buffer, with the additional time-spanning nature of exponential decay.
This representation is concatenated, flattened, and passed to the agent.
The resulting exponentially-decaying feature set is illustrated in
Figure \ref{fig:rep_viz}C. We tested agents given this representation on
Catch with logarithmically spaced mask sizes of 0, 1, 2, 4, 8, and 16.
As we will explain below, understanding the results of these tests was
aided by testing on a slightly modified environment. These similar
tests, with agents given this same exponentially-decayed feature
representation, were tested on Catch with only 1 ball per game; the
environment shape and mask sizes were kept to be the same as previous
tests.

Finally, we also trained and tested an agent using our SITH
representation with five nodes per feature. An illustration of the input
layer during a moment of gameplay with a SITH representation of 5 nodes
is shown in Figure \ref{fig:rep_viz}A. Note the ``smearing'' of the
temporal representation and also how the nodes span logarithmic (as
opposed to linear) distances into the past. Each agent was then tested
on versions of Hidden Catch with logarithmically-spaced mask sizes of 0,
1, 2, 4, 8, and 16.

\begin{figure}
\centering
\includegraphics[width=1\textwidth,height=\textheight]{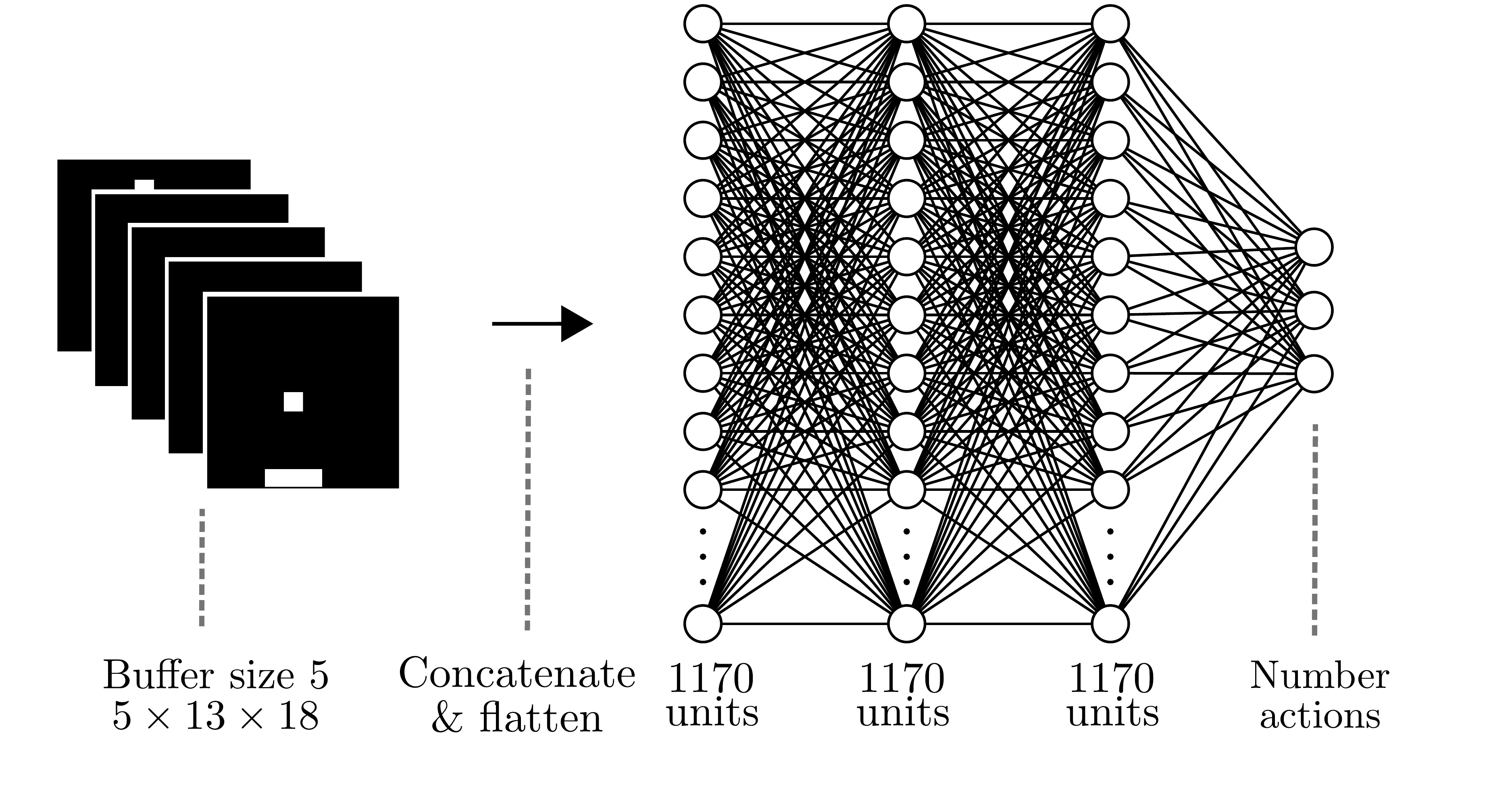}
\caption{\emph{Fully connected DQN with representation outside the
network allows for comparisons between working memory input
representations.} The most recent frame from the catch game is input
into a FIFO buffer. Then we concatenate the frames and flatten into a
\(1 \times 1170\) (\(5 \times 18 \times 13\)) vector. That vector is the
input to our DQN, which consists of 2 fully connected hidden layers,
same size as the input vector, and outputs a \(1 \times 3\) vector where
the max of those 3 numbers indicates the action our agent should take
that frame.\label{fig:model}}
\end{figure}

During our simulations with Catch, we used simple densely-connected
feedforward networks, where the number of parameters depended on the
number of indices \(N\) used in the representation model. All networks
trained on the Catch game contained 4 layers: an input layer of size
\(S \times N\), two hidden layers of the same size as the input layer,
and an output layer of size 3, to indicate all possible movements of the
basket (left, right, and no-op). An illustration of the model with a
FIFO buffer of size 5 is shown in Figure \ref{fig:model}.

All densely-connected networks learned to approximate the Q function in
a Q-Learning RL paradigm, with a discount rate of 0.9. The memory store
for experiencing replay retained up to 50 previous full games, with each
game consisting of 10 balls worth of frames. For each frame observed
from the environment, the network went through one session of
experiencing replay, which consisted of replaying 10 random full games
from its memory store. These networks were trained using the Adagrad
optimizer with a learning rate of 0.01, an epsilon of
\(1 \times 10e^{-8}\), and decay of 0 (Chollet, 2015/2015). Networks
were trained and tested on 500 epochs, where an epoch consisted of the
back-propagation of error from a session of experiencing replay.
Networks were tested on 100 games every 5 epochs of training, and tested
on 1000 games after completing training. These models were trained and
tested during five independent learning runs to quantify variability in
learning performance. We note here that for our tests on Catch with with
one ball per game, each game was \(1/10^{\text th}\) the length of other
tests with Catch. For a more even comparison to other results, the
replay buffer was increased to a size of 500 games, and each network was
tested for 5000 epochs. This ensured that each network was able to
replay experiences similar distances into the past, and was trained on a
similar number of frames.

When testing the models that received either SITH or exponentially
decaying representations as the input, we had to make some assumptions
with regard to real-world timing. The first was to treat each Catch game
frame as if \(1/30\) seconds passed. This assumption allowed these
representations to update their array of decaying cells in a real unit
of time. Specifically, each frame of Catch was presented for
\(1/300^{\text th}\) of a second and decayed for \(1/30 - 1/300\)
seconds.

For the SITH model, due to the overlap in the receptive fields, there
was no need to pass in the the entirety of
\(\tilde{f}(\overset{*}{\tau})\), for every feature, into the neural
network. Instead, we ``sliced'' into SITH at logarithmically-spaced
points into the past. Here, the set of all temporal receptive fields
were centered around
\(\overset{*}{\tau} = [0.03333333, 0.07145296, 0.15316577, 0.32832442, 0.70379256]\),
which are in units of seconds. The size and shape of these receptive
fields are shown in Figure \ref{fig:rep_activation}A. In effect, SITH
acts as a ``blurred'' representation of the past, where each ``slice''
into the past was a temporally-blurred game screen of features, each
containing varying amounts of activation and temporal specificity.

Similar to the buffer, both the exponentially decayed and SITH
representations were concatenated and flattened before acting as input
to the network. The 5 time indices pulled from SITH gave rise to network
parameters equal in size to that of the model with a buffer size of 5;
i.e., 1170 input features. This was also the case with the 5
exponentially decayed feature sets. Networks that are given either an
exponentially decayed representation or SITH follow the exact
configuration as in Figure \ref{fig:model}, but with a corresponding
input representation in place of a FIFO buffer.

\hypertarget{models-for-playing-flappy-bird}{%
\subsection{Models for Playing Flappy
Bird}\label{models-for-playing-flappy-bird}}

\begin{figure}
\centering
\includegraphics[width=1\textwidth,height=\textheight]{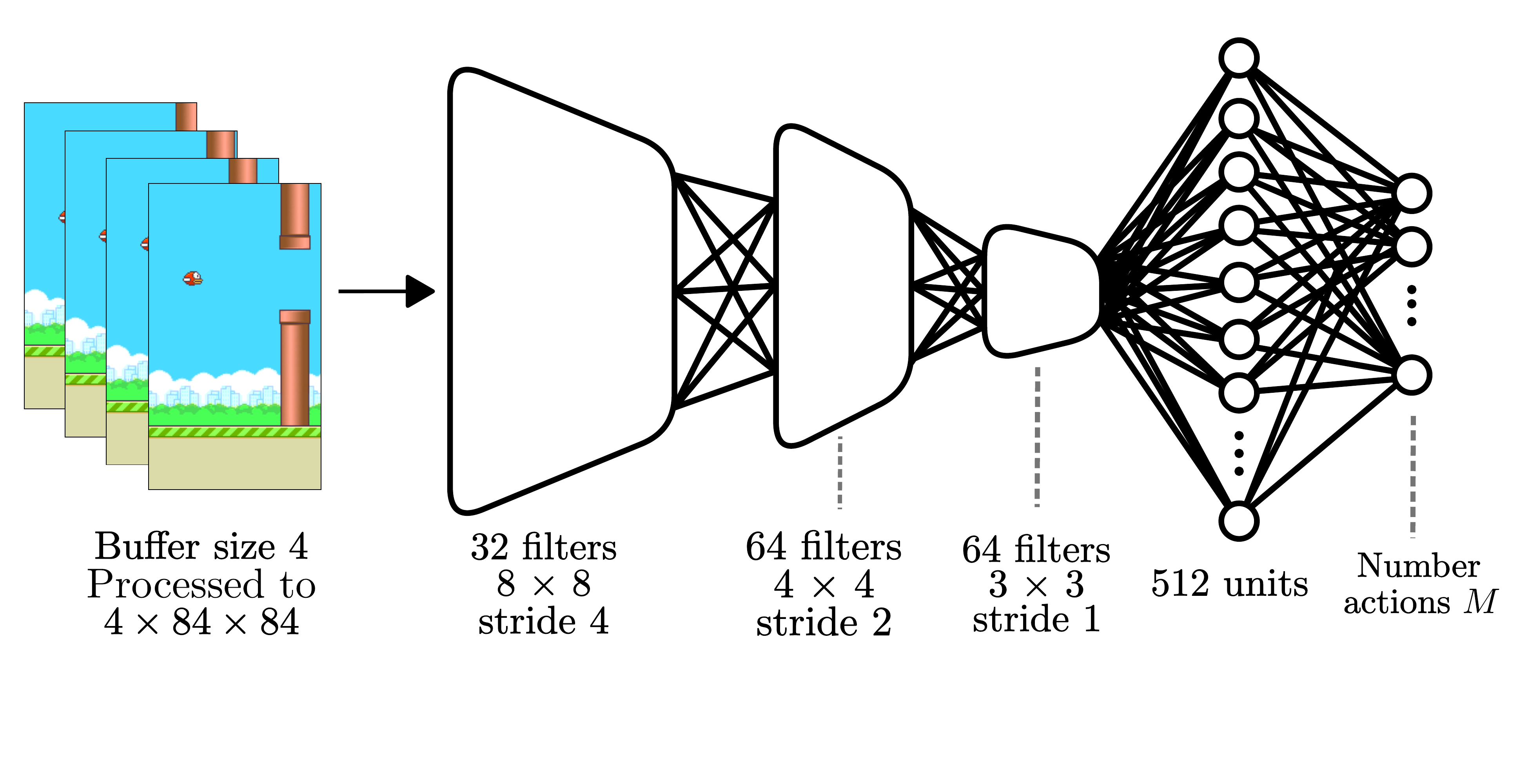}
\caption{\emph{A standard DQN architecture with convolutional layers
allows comparisons between buffer and SITH representations in
more-complex games.} Game frames are preprocessed in a manner described
in the main text, and illustrated in Figure \ref{fig:flappy_bird}B, C,
E, and F. The various representations are processed through 3
convolutional layers, then passed through one fully connected hidden
layer, to the output layer. In Flappy Bird, the number of actions
\(M = 2\). Each output unit predicts the future reward for a given
action, and the maximum rewarded action is chosen for the agent to
perform. Only a buffer representation is illustrated here, but a SITH
representation utilizes the same network design and frame
preprocessing.\label{fig:conv_net}}
\end{figure}

After testing our models in the game Catch, we sought to test SITH in a
noisier, more compelling environment, requiring a more specialized
network architecture to learn the task. For this comparison, we measured
the performance of a buffer of size 4 against a SITH representation with
4 ``slices'' into the past. Here, SITH is indexed at
\(\overset{*}{\tau} = [0.03333334, 0.08645808, 0.22425, 0.58164674]\).
As we did with the game Catch, we set the length of one frame to be
\(1/30^{\text th}\) of a second; the input screen was presented for
\(1/300^{\text th}\) of a second, followed by a delay of
\(1/30 - 1/300\) seconds.

Taking after Mnih et al. (2015), we utilized a convolutional neural
network (CNN), illustrated in Figure \ref{fig:conv_net}, with the same
structure (Jo, 2017/2017). In the preprocessing step, the
representations are converted to grayscale, then down-sampled and
trimmed to a size of \(84 \times 84\) pixels. This
\(4 \times 84 \times 84\) array of feature sets is then given to the
network as input. An illustration of this step is shown for a buffer in
Figure \ref{fig:flappy_bird}C and for SITH in Figure
\ref{fig:flappy_bird}B. The same preprocessing was performed on the
partially obscured variant of Flappy Bird, as seen in Figure
\ref{fig:flappy_bird}E and F.

The network was trained with the Adam optimization algorithm, with a
learning rate of 0.0001, and a smooth L1 loss function to calculate
error (Paszke, 2016/2017). Training occurred over 1,000,000 epochs,
where each epoch was one frame of Flappy Bird. Training was performed in
5 independent runs for each representation (buffer and SITH), and for
each game variant (fully visible and partially obscured Flappy Bird),
resulting in 4 interactions being tested. During each epoch, a random
minibatch of 32 previous \((s_t, a, s_{t+1}, r)\) (state, action, next
state, reward) tuples was pulled from the experience replay buffer,
which itself had a maximum size of 50,000 tuples. This minibatch was
used to backpropagate errors through the network, with one minibatch for
every epoch. By the means of a DQN, this network approximated the
Q-function found in Q Learning, with the discount factor
\(\gamma = 0.99\). An \(\epsilon\)-greedy policy was utilized to balance
exploration and exploitation, where \(\epsilon\) started at 1.0, and was
exponentially reduced to 0.1 according to
\(\epsilon_{i} = \epsilon_{i-1} e^{\frac{\text{epoch}_i}{\text{decay}}}\),
with a decay of 100,000. Similar to Mnih et al. (2015), a target network
had its weights periodically copied from the network being trained, and
was used to calculate \(Q(s_{t+1}, a_{t+1})\) when determining loss.
This target network was updated every 1000 epochs, and increased
training stability in the ``true'' network. During training, one test
game of flappy bird was played every 900 epochs, giving a view of how
the network performed while training.

\hypertarget{results}{%
\section{Results}\label{results}}

\hypertarget{performance-in-catch}{%
\subsection{Performance in Catch}\label{performance-in-catch}}

\begin{figure}
\centering
\includegraphics[width=0.65\textwidth,height=\textheight]{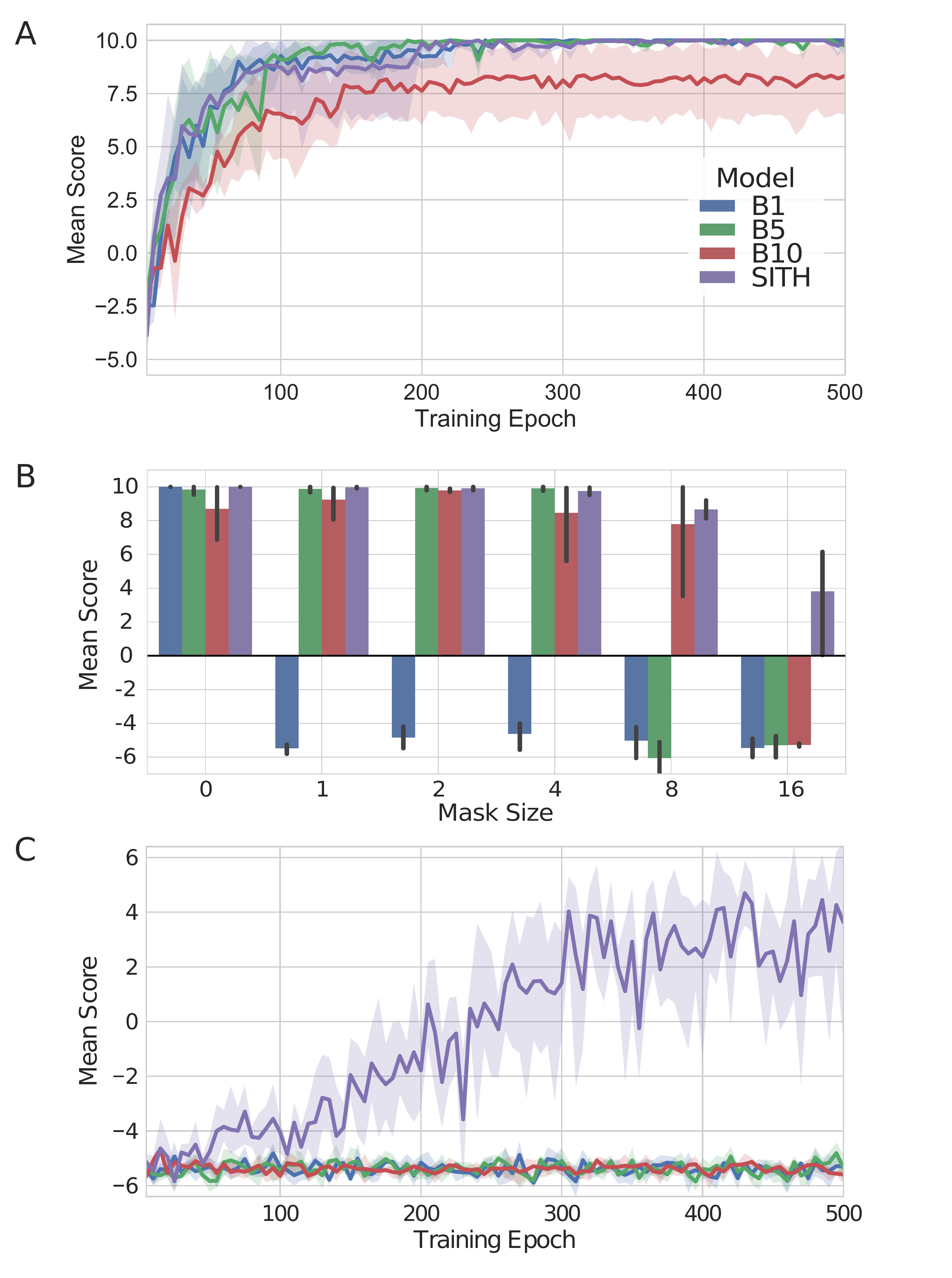}
\caption{\emph{Buffers fail when the ball cannot be seen at reward time
while SITH performs the same or better than a buffer for all mask sizes.
} (A) Performance of SITH and buffers in fully visible Catch. With no
obfuscation, both SITH and a FIFO buffer are sufficient for learning
Catch. This learning occurs at a similar rate, given the same network
architecture and training parameters. Shaded regions indicate 90\%
confidence interval over 5 independent simulations for each model, with
each model tested on 100 games at each point. (B) Trained performance of
each model over mask sizes. As the size of the mask (i.e.~amount of
feature obfuscation) increases, buffers fail when rewarding information
is outside the buffer size. Meanwhile, SITH is able to match buffer
performance in lower mask sizes, while maintaining rewarding information
over longer timescales. This leads to performance well above chance in
all configurations. Plotted here is mean performance on 1000 games for
each model after training, with 5 independent simulations per model and
per mask size. Black lines indicate the 95\% confidence interval about
the mean. (C) Performance of SITH and buffers in fully obscured Catch.
On a mask size of 16 (the max amount of obfuscation with this
configuration) buffers of sizes 1, 5, and 10 all fail to learn even the
relatively simple Catch game. SITH is sufficient for performance well
above chance, due to its farther-reaching temporal span. Shaded regions
indicate 90\% confidence interval over 5 independent simulations for
each model, with every model tested on 100 games at each
point.\label{fig:catch_perf}}
\end{figure}

Agents given either a buffer or SITH as representations were trained on
the fully visible catch game. As illustrated in Figure
\ref{fig:catch_perf}A, these models all performed well; over 1000
testing games, they received an average score at or close to 10, the
maximum score possible. Despite having more information about the
environment than a FIFO buffer of 1, the models based on FIFO buffers of
size 5 and 10 had some variance in their final performance. This was
most likely due to the corresponding networks having many parameters
with extraneous information leading to overfitting. Overall, high
performance was to be expected and served to establish a baseline for
the critical test offered by Hidden Catch.

After establishing baseline functionality by training and testing on the
fully visible game of Catch, we then trained and tested agents with
these representation configurations on Hidden Catch. These trials were
run using logarithmically-spaced mask sizes, as seen in Figure
\ref{fig:catch_perf}B. This illustration demonstrates the inflexible
nature of the FIFO buffer. Each network performed well above chance as
long as its FIFO buffer was able to capture the ball before being hidden
behind the mask. However, as soon as the mask obfuscated that
information from the network's history, the network failed. This is
because the models that use FIFO buffers are not able to bridge temporal
gaps larger than the size of the FIFO buffer. The ball is hidden from
the network during the most critical observations. For example, the FIFO
buffer of 1 could only present the most recent observation to the
network, but performed optimally in a fully visible environment. A
similar result occurs when testing the FIFO buffer of size 5 with a mask
of 8, and a FIFO buffer of size 10 on a mask of 16.

We note the success of the model that took a SITH buffer as input at a
wide range of time scales, also seen in Figure \ref{fig:catch_perf}B.
The agent learned to play all versions of Hidden Catch with scores well
above chance due to the logarithmic spacing and preservation of time.
Even in the most difficult scenario, where all but two rows were hidden
from the network, the SITH model performed well above chance, as seen in
Figure \ref{fig:catch_perf}C. Thus, the network is able to create a
mapping of the ball's location with that of a reward. As is the nature
of SITH, however, this lossy representation is also responsible for
slower learning during training with increasing mask sizes, as well as
higher variance in the final scoring benchmark; this is visualized and
most clearly evident for the mask sizes of 8 and 16 in Figure
\ref{fig:catch_perf}B.

\begin{figure}
\centering
\includegraphics[width=0.8\textwidth,height=\textheight]{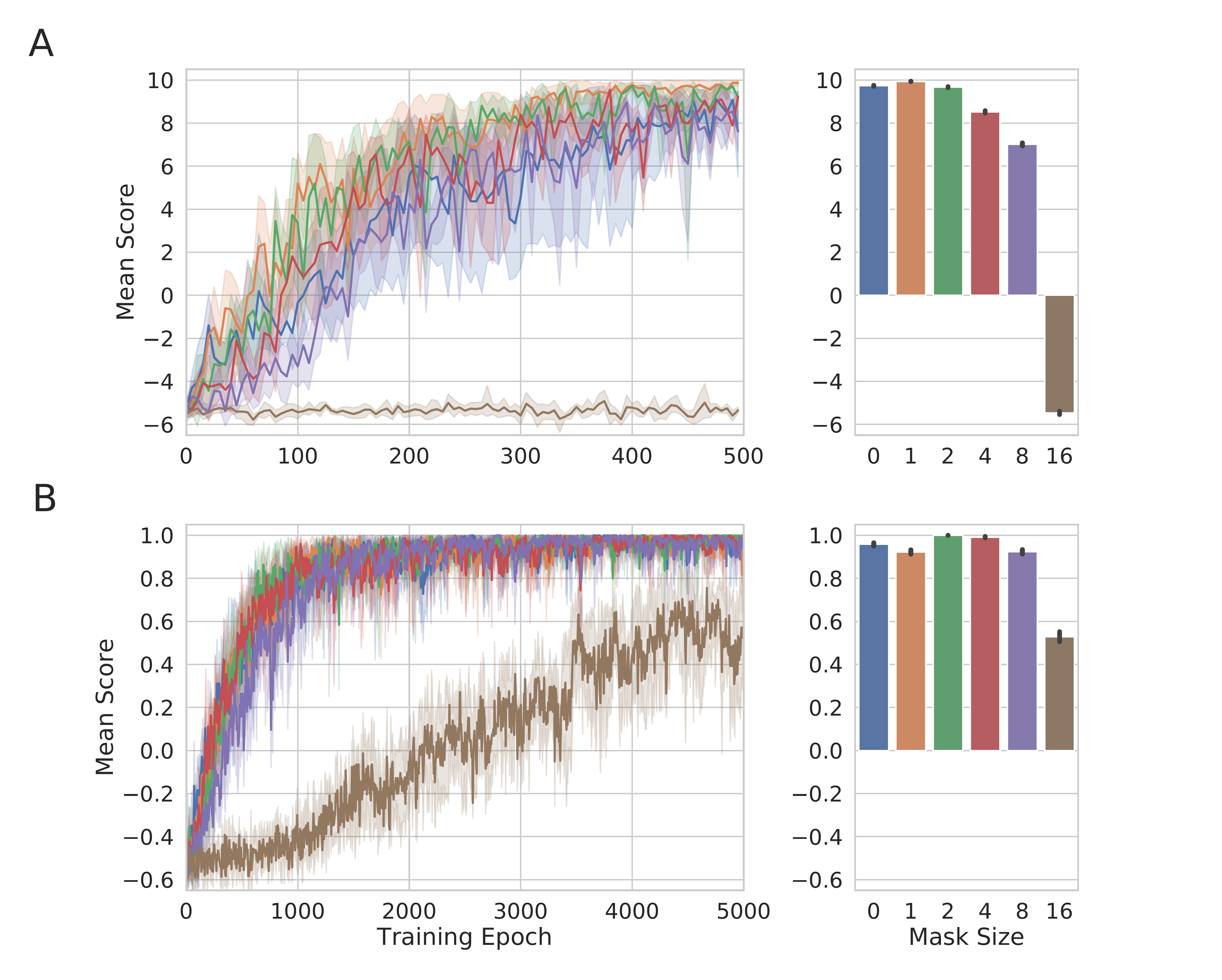}
\caption{\emph{Exponentially decayed sets do not clearly encode past
events.} All shaded regions and bar extensions indicate 95\% confidence
intervals. (A) Performance of exponential decay on Catch with 10 balls
per game. Performance as a function of training epoch is shown, with 5
independent simulations run for every game configuration. The
representation has the necessary temporal reach for all levels of
obfuscation. However, with 10 balls in every game, noise from previous
balls hampers learning, especially with high obfuscation. Also plotted
is the final performance for each game type, after training on 1000
games. (B) Performance of exponential decay with 1 ball per game.
Performance is shown during training, for 5 independent simulations per
game configuration. By reducing the number of balls per game, no noise
is carried over from previous balls. The representation has sufficient
temporal span, and becomes more separable with the reduced noise. Final
performance of each simulation and game configuration, over 1000 games,
is plotted on the right.\label{fig:little_t_perf}}
\end{figure}

To further understand the potential advantages of SITH, we also tested
agents with an array of exponentially decaying feature sets on Catch and
Hidden Catch. The structure of this representation is illustrated in
Figure \ref{fig:rep_viz}C. As expected, this model performed nearly
optimally in the visible variant of this game, as seen in Figure
\ref{fig:little_t_perf}A. However, it performed no better than chance
when tested on Hidden Catch with a mask size of 16. This is in contrast
to the performance of SITH on the same game, which performed well.

Due to the selected values of \(s\), this representation spanned further
back into the past than did the SITH representation. This suggests that
when representing the past, more is needed than simply spanning longer
durations. Supporting this hypothesis demanded further testing with
exponentially-decaying sets of features. To simplify the task, the state
space of the exponentially-decaying set had to be reduced. This was
achieved by providing only 1 ball for every game of Catch, causing a
reset of the representation after each ball hit the ground. By not
maintaining irrelevant information on previous balls, the number of
possible values of the representation decreased. Testing was again
performed on logarithmically-increasing mask sizes, and results are
shown in Figure \ref{fig:little_t_perf}B. While the agent took
significantly more epochs to learn the reward structure with a mask of
16, the representation is clearly capable of capturing long-term
dependencies, otherwise the agent would never perform above chance.

\begin{figure}
\centering
\includegraphics[width=1\textwidth,height=\textheight]{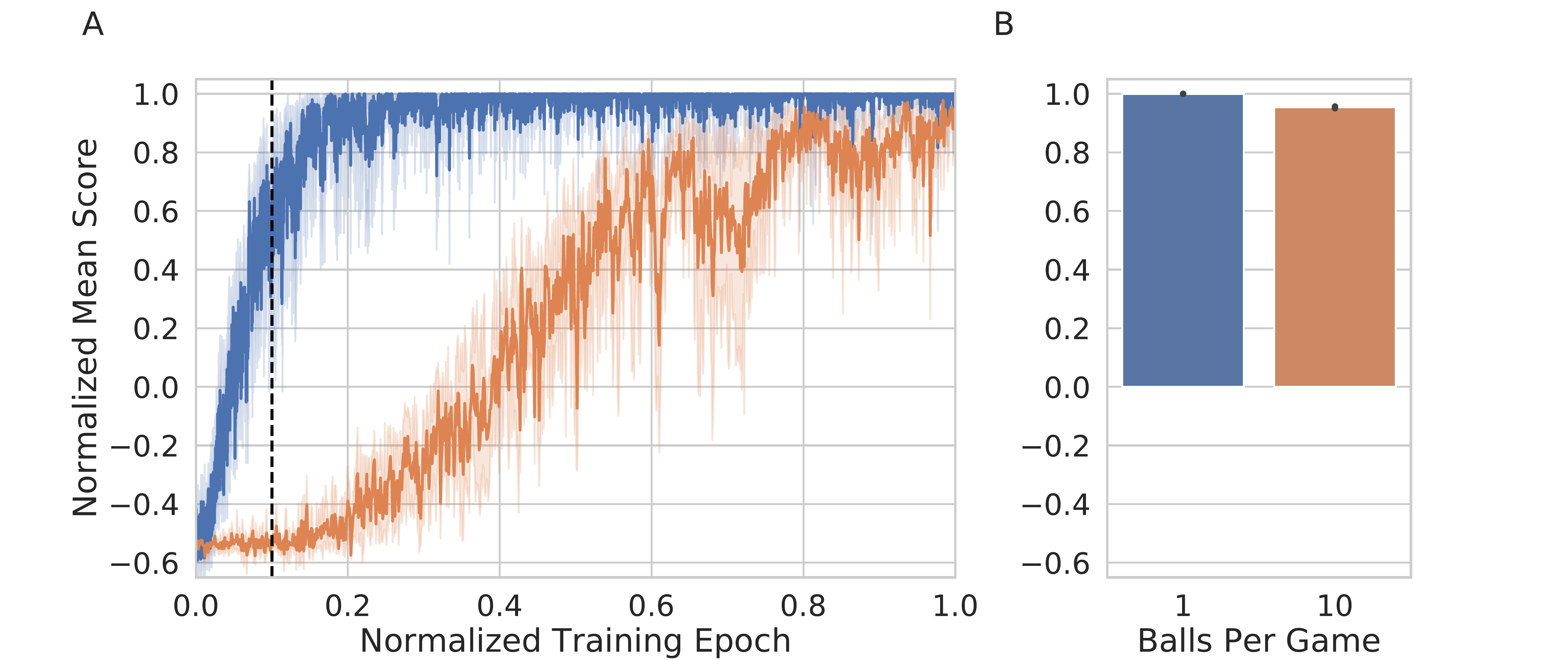}
\caption{\emph{Exponentially-decayed sets require five times more
training than SITH on the same Catch parameters.} With
exponentially-decaying feature sets, networks are able to achieve near
optimality given longer training times. Scales are normalized for
clarity of comparison. In order to equate the total amount of training,
networks trained with 1 ball per game were trained for 50000 epochs,
with chance performance at -0.5 and possible mean scores between -1 and
1, inclusive; networks trained with 10 balls per game were trained for
5000 epochs, with chance performance at -5 and possible mean scores
between -10 and 10, inclusive. (A) Performance of networks during
training. Each line corresponds to network performance while being
trained on games with different numbers of balls per game; the dashed
vertical line indicates where training was stopped in previous tests.
Shaded regions indicate 95\% confidence intervals. The number of balls
per game has a significant effect on performance, though the information
needed for optimal performance is still present in the
exponentially-decaying set. Nevertheless, agents with the
exponentially-decayed representation required 5 times as many epochs to
match the performance of agents with SITH, as seen in Figure
\ref{fig:catch_perf}C. (B) Testing performance of networks trained on
each game after training; error bars indicate 95\% confidence interval.
Despite the slower rates of learning, networks with
exponentially-decaying sets tested in both environments were capable of
achieving near optimality.\label{fig:long_little_t}}
\end{figure}

Knowing this capability, agents with exponentially-decaying sets were
trained on games with a mask of 16, and either 1 or 10 balls per game,
and were trained an order of magnitude more epochs. Results of this more
focused testing are shown in Figure \ref{fig:long_little_t}. Here, it is
shown that the reward structure of both game configurations can be
learned, given enough training epochs. We note that the exponentially
decaying set required a fivefold increase in epochs to match the
performance of SITH on the same task (as seen in Figure
\ref{fig:catch_perf}C), and a tenfold increase to achieve near
optimality.

\hypertarget{performance-in-flappy-bird}{%
\subsection{Performance in Flappy
Bird}\label{performance-in-flappy-bird}}

\begin{figure}
\centering
\includegraphics[width=0.8\textwidth,height=\textheight]{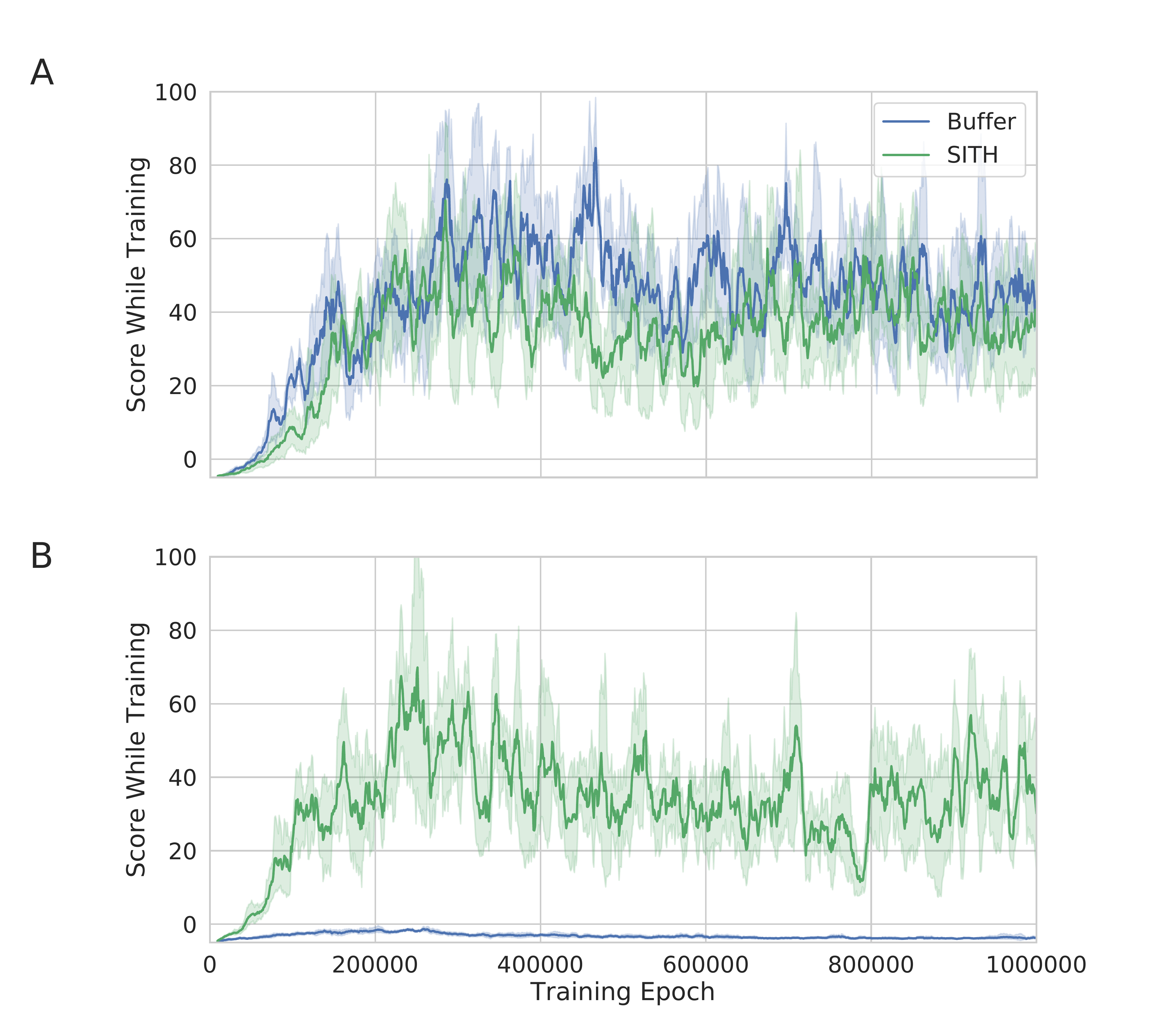}
\caption{\emph{SITH performance matches or exceeds a buffer, even with a
CNN required to learn a more-complex game.} All shaded areas indicate
95\% confidence intervals about the mean. For clarity, all plots are
shown with a rolling mean with a window size over 10 epochs. (A)
Performance on regular version of Flappy Bird. Without any obfuscation,
both the FIFO buffer and SITH are capable representations for learning
Flappy Bird. Shown here are 5 independent simulations for each model.
(B) Performance on partially obscured Flappy Bird. By obscuring pipes in
game, the buffer is unable to maintain information for rewarding
features. On the other hand, SITH is able to maintain this information,
even when passed to a convolutional layer. Shown are 5 independent
simulations for each model.\label{fig:flappy_perf}}
\end{figure}

After demonstrating the advantage of SITH over other models, we wanted
to measure performance of SITH in a more ``real-world'' setting,
requiring more complicated neural network models to attain high levels
of performance. As outlined above, we selected a variant of the game
Flappy Bird, which requires training a convolutional neural network to
solve it. Our first simulation was to establish baseline performance and
verify that both the buffer and SITH models could learn to play a fully
visible variant of Flappy Bird. As shown in Figure
\ref{fig:flappy_perf}A, the performance of both models is comparable,
with overlapping 95\% confidence intervals over the independent training
and testing runs.

When testing in the partially observable variant of Flappy Bird, results
are consistent with those in the Catch game. As seen in Figure
\ref{fig:flappy_perf}B, a SITH representation allows the network to
achieve scores similar to those of the fully visible simulations. For
the buffer representation, however, there is no discernable pattern of
learning and the network rarely performs above chance. These results
demonstrate that SITH can be used as input to a convolutional neural
network. This network can then extract meaningful spatio-temporal
representations that can help solve partially-observable complex
environments which require representations of the past to guide actions
in the present.

\hypertarget{discussion}{%
\section{Discussion}\label{discussion}}

\begin{figure}
\centering
\includegraphics[width=0.5\textwidth,height=\textheight]{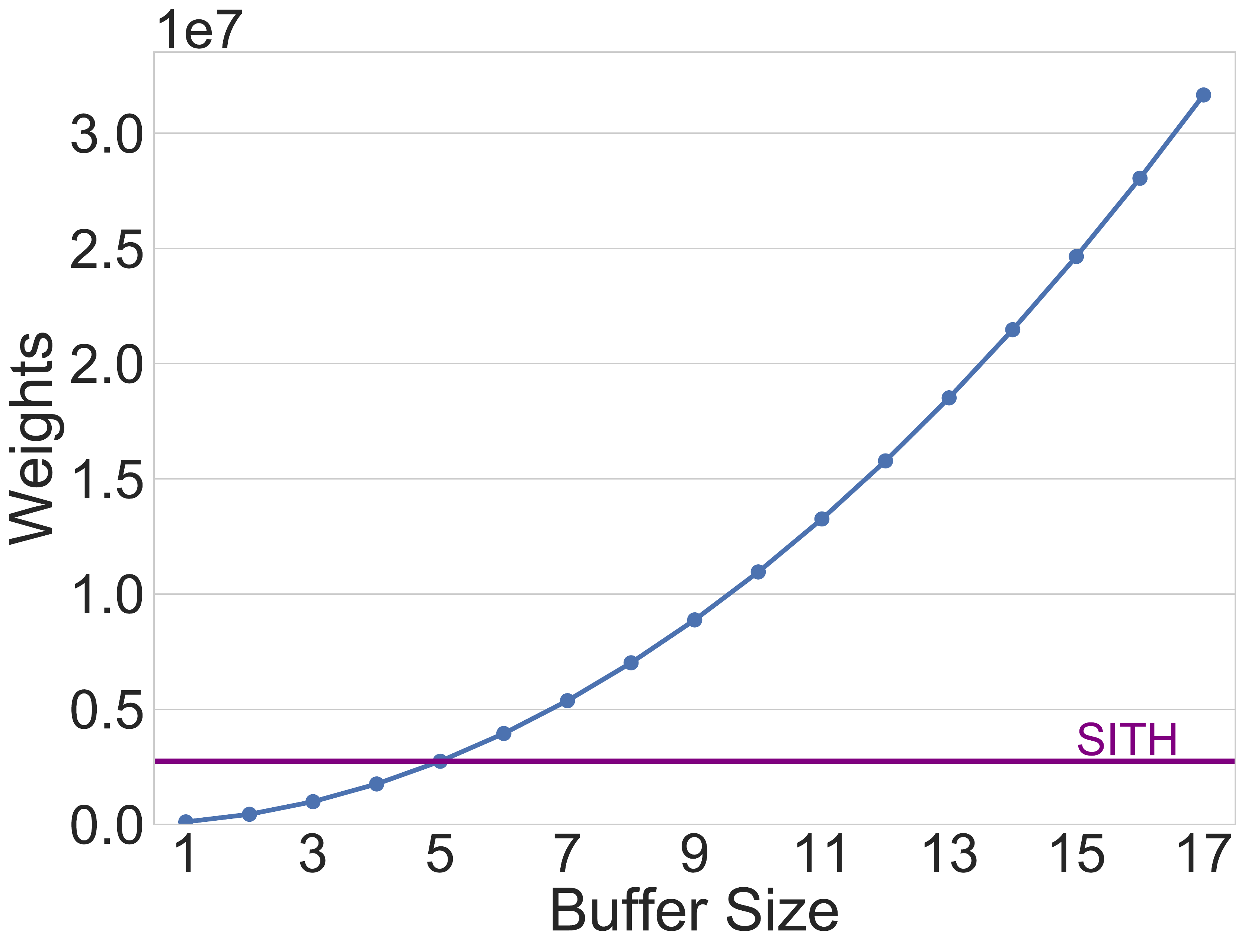}
\caption{\emph{SITH conserves resources by keeping a compressed
representation of the past.} With two fully-connected hidden layers in
our model, the number of weights needed to be learned grows
quadratically with the size of the buffer. In order to solve the hidden
catch game with a mask size of \(N\), we would need a FIFO buffer of
size \(N + 1\). The horizontal line indicates the number of weights for
the model using SITH with 5 nodes, which was able the solve the hidden
catch game up to a mask size of 16.\label{fig:num_parameters}}
\end{figure}

We presented a novel approach for representing the history of features
in a scale-free, optimally fuzzy, and neurally-plausible manner. We
tested a DQN using a SITH representation as input for a toy example of
the game Catch, and its variant, Hidden Catch. We verified that the
model, given SITH as input, performs well when features are fully
visible, and, more importantly, when features are partially observable.
Additionally, we showed that this performance in Hidden Catch only needs
the same number of model parameters as a model based on a FIFO buffer of
size 5 (Figure \ref{fig:num_parameters}) and still outperforms models
with a FIFO buffer of size 10 across a wider range of mask sizes.
Moreover, we determined that a FIFO buffer of size 17 would be needed to
perform well in a game of fully hidden catch at the cost of many more
resources than a SITH representation with 5 nodes. We also tested a set
of exponentially decaying features in the same environment and found it
required many times the number of training epochs for a highly obscured
task. This test suggests that temporal reach is not the only important
feature of a representation, but also clarity and information of what
happened when.

Finally, we also tested our model in a more complex, noisy environment,
using more specialized CNNs. As it was in a simple environment, SITH was
a capable representation in both fully visible and partially observable
environments, while a FIFO buffer failed in the partially observable
variant. The successful utilization of CNNs, and its ability to track
visual features even in SITH's blurred representation, further
demonstrated SITH's utility as a drop-in replacement for a FIFO buffer.

\hypertarget{interpretation-of-results}{%
\subsection{Interpretation of Results}\label{interpretation-of-results}}

When viewed from the lens of the problem domain, at least two reasons
can explain SITH's performance, and the buffer's malperformance. To
provide a formal description of these environments, consider a Markov
Decision Process (MDP). According to Kaelbling, Littman, \& Cassandra
(1998), an MDP is a model of a domain in which the current observable
state is sufficient in determining the optimal action for transitioning
to subsequent states. Here, ``optimal'' is defined as that choice that
maximizes some future reward inherent in the domain. While the
constraints of an MDP do not require the agent to process information
outside the current state, a Partially Observable MDP (POMDP), where
there is uncertainty in the state observations, can require such
information. Here, the exact status of a state ranges over some
probability distribution, and, consequently, the agent must navigate an
uncertain world. One approach to optimizing performance in such
environments is to store previous states. All three representations
tested here store and transform a history of previous states, letting
the agent find rewarding correlations between states. By taking
advantage of this correlation through time, the agent can better
determine the environment's current state, giving it a higher
probability of selecting actions that will maximize future reward.

The fully visible game of catch is an MDP, as shown by the optimal
performance of a FIFO buffer of size 1. However, once a mask is
introduced, the domain becomes a POMDP where states correlate in time.
By maintaining a far-reaching representation of the past, SITH allows
the agent to reduce the size of the probability space for the current
state. The construction of SITH increases this temporal span with only
\(log \ N\) memory resources, while a FIFO buffer would require \(N\)
resources. This lets the agent make a more informed decision about its
next action with a more tractable number of computations, as long as
SITH is parameterized to cover the necessary temporal distance.

But temporal reach cannot be the only cause of this performance. As
parameterized in these tests, the exponentially-decaying feature set
spans farther in time than SITH. Despite this, agents with this
representation required significantly more training epochs (Figure
\ref{fig:long_little_t}A). This can be explained by the temporal
specificity found in SITH and the FIFO buffer relative to the
exponentially-decaying representation. As illustrated in Figure
\ref{fig:rep_activation}A, C and Figure \ref{fig:rep_viz}A, C, the
receptive fields in SITH focus on \emph{specific} temporal ranges in the
past. While the specificity becomes logarithmically blurred
(i.e.~compressed) as it extends further into the past, the recreated
timeline still centers on a specific point. This method of history
recreation can be more computationally useful, as it weighs activations
so an agent can more clearly determine when that feature was active. Put
more succinctly, SITH not only provides a representation \emph{that}
something happened, but also \emph{when} it happened, providing an
explanation for why an exponentially-decaying representation required a
fivefold increase in training epochs relative to SITH.

Given that both the \(\mathbf{L}^{-1}_k\) operator in SITH and
fully-connected neural networks perform linear combinations of
activations, it is important to note that a deep neural network could
learn to approximate the inverse Laplace transform given a large-enough
set of decaying representations and enough training. For example, with
the simulations and model parameters presented here, specifically the
value of \(k=4\) for the Post approximation of the inverse Laplace
transform, this would mean we would need a family of 9
exponentially-decaying representations for each of the 5 SITH
timescales, so 9-times as many features as input into the model. Thus,
similar to the resource requirements in a FIFO buffer, a neural network
could emulate the advantages provided by SITH, but at a cost. The
required training epochs for such a large number of features (9 times as
many as SITH) would likely be wasteful, especially when compared to the
numerically well-defined values of \(\mathbf{L}^{-1}_k\).

\hypertarget{sith-in-relation-to-previous-models}{%
\subsection{SITH in Relation to Previous
Models}\label{sith-in-relation-to-previous-models}}

SITH fits well alongside previous work on delayed signals in RL models.
The exponentially decaying feature representations, as used in our
second set of experiments, are similar to models described in Stephen
Grossberg \& Schmajuk (1989) and Kurth-Nelson \& Redish (2009), as well
as those reviewed in Langdon, Sharpe, Schoenbaum, \& Niv (2018). Indeed,
SITH relies on these decayed representations to perform the Laplace
transform on its input. However, SITH then performs local integration,
via the inverse Laplace transform, on these decayed representations,
which is ultimately how SITH maintains a compressed history. As shown in
our second experiment, this integration results in a qualitatively
different representation that is better able to capture how long ago in
the past a feature was active. This approach more closely resembles
models such as those in S. Grossberg \& Merrill (1992) and Ludvig,
Sutton, \& Kehoe (2008). Critically, though, these models do not
maintain scale-invariance, placing SITH in a unique position among
previous approaches.

It is pertinent to mention SITH's capabilities in relation to other
methods utilized by neural networks for maintaining features through
time. One of the more prominent is the Long Short-Term Memory (LSTM)
architecture (Hochreiter \& Schmidhuber, 1997). While both SITH and
LSTM's seek to capture temporal dependencies in uncertain environments,
it is difficult to compare these approaches directly because they have
very different properties. Whereas LSTM's can learn to maintain features
for long durations (by means of logic gates to store and erase
information), SITH effectively provides a logarithmically-spaced
spectrum of decay rates that follow Weber-Fechner scaling laws seen in a
range of behavioral and neurophysiological domains with no learning
required. Whereas LSTM events can be triggered by external events and/or
internal states, SITH provides a veridical (although coarse-grained)
record of the past. For these reasons it is not straightforward to
compare the resources or learnability associated with each method.
Moreover, SITH must be coupled to some form of learning architecture to
be useful for actual applications. That said, future work will entail
more detailed tests of whether there are cases in which SITH or LSTM
models can outperform, or even complement, each other on various tasks.

More recently developed methods, such as the read-write memory matrix
found in Neural Turing Machines (Graves, Wayne, \& Danihelka, 2014) and
the Differentiable Neural Computer (Graves et al., 2016) also warrant
mention. As with LSTM's, such methods are computationally useful, but
are in a separate class from SITH. Through the adjustment of read-write
weights through gradient descent, such methods integrate the memory
representation into the agent itself, dynamically adjusting its history
based on learned utility. In contrast, SITH separates the parameters of
the representation from the parameters of the agent, where the agent
cannot tune the representation parameters, just how it weighs those
features internally. However, as with LSTM's, comparing and combining
SITH with such models may be fruitful for future work.

\hypertarget{future-work}{%
\subsection{Future Work}\label{future-work}}

In general, due to its flexibility and efficiency, the SITH model
provides a myriad of opportunities for future work. As mentioned before,
the Catch and Flappy Bird games provide examples of SITH supplying the
agent with information necessary for optimal decision making in an
obscured environment. It is reasonable to expect SITH to be a similar
aid in other environments that contain temporal dependencies at various
scales. Considering the breadth of time covered by a SITH
representation, we would expect our model to aid in many different
environments with minimal parameter adjustment.

One property not utilized in these experiments is the fractional,
fine-tuning of scale in SITH. While varying mask sizes does entail a
change in scale (as the agent must maintain information about the ball
for differing lengths of time), this modulation is rather discrete, and
coarse. A more fine-grained tuning of SITH would conceivably allow any
SITH-enhanced agent to adapt on the fly to arbitrary rate changes in an
environment, simply by modifying a single parameter \(\alpha\), which
scales the rate of change in the SITH representation (K. H. Shankar \&
Howard, 2012). For example, imagine learning to play a game well at one
speed and then having to play it at another, previously unseen, rate.
This is achievable, within reasonable bounds, by an average human agent.
However, many AI representations of the past would have their structure
altered by a change in scale; this leaves the agent vulnerable to
performance drops without any mechanism to adjust to this change without
significant retraining. SITH instantiates such a mechanism, and could
allow for this more human-like behavior in artificial agents. Moreover,
tuning this single SITH parameter could, in principle, be learned by the
agent itself, minimizing the amount of parameter tuning for any given
agent. This is similar to Sozou (1998), where a ``hazard rate'' (the
rate at which a reward becomes increasingly unlikely) was automatically
updated based on inconsistencies in reward dispensation. One could
imagine using a similar method to automatically tune \(\alpha\) based on
irregularities in perceived, or even predicted, environment states.

As discussed previously, a learning architecture that commits to a
particular scale is intrinsically limited in its flexibility. The size
of the FIFO buffer \(N\) fixes a scale. In addition, the factor
\(\gamma\) in the Bellman Equation for Q-learning also introduces a
scale. The Bellman equation efficiently implements exponential
discounting where outcomes \(\delta\) steps into the future are
discounted by \(\gamma^\delta\). This exponential discounting introduces
a scale as the generalization across time is very different for
\(\delta \gg \log \gamma\) and for \(\delta \ll \log \gamma\). Future
work in RL should endeavor to replace the scale introduced by a fixed
\(\gamma\) in Q-learning. One possibility is to replace a single
\(\gamma\) with an ensemble of Q-learning implementations with different
learning rates---a similar approach was taken to Temporal Difference
learning in Kurth-Nelson \& Redish (2009). One could also note that an
ensemble of Q-learning models with a spectrum of values of \(\gamma\)
encodes the Laplace transform over future time points and use the Post
approximation to invert the Laplace transform, resulting in a compressed
estimate of future outcomes. This is very much in keeping with the
spirit of the SITH method itself.

One final possibility is to replace exponential discounting entirely
with a power law-scaled prediction of the future, as shown in Z. Tiganj
et al. (in press). Through the use of a single linear operator on a SITH
representation, one would retain all advantages of the SITH model, but
with an estimated value of the future. This future representation would
remain compressed, with the estimation of events increasing like
\(log \ N\) while predicting \(N\) points into the future. This
representation would, like SITH, remain scale-invariant, where the
predicted value of a future state continues to be scaled by the same
\(\alpha\) as described above. Such a model would allow for prediction
of rewards over multiple scales and could greatly improve reinforcement
learning algorithms that rely upon exponentially discounted
expectations. With a SITH representation of both the past and the
predicted future, AI can acquire a human-inspired sense of time.

\hypertarget{acknowledgments}{%
\section{Acknowledgments}\label{acknowledgments}}

We thank Zoran Tiganj for insightful discussions and the anonymous
reviewers for helpful feedback on earlier versions of this work. This
work was supported by National Science Foundation (NSF) grants 1631403
(PBS) and 1631460 (MWH).

\hypertarget{references}{%
\section*{References}\label{references}}
\addcontentsline{toc}{section}{References}

\hypertarget{refs}{}
\leavevmode\hypertarget{ref-Ba.etal.2016}{}%
Ba, J., Hinton, G., Mnih, V., Leibo, J. Z., \& Ionescu, C. (2016). Using
Fast Weights to Attend to the Recent Past. In \emph{Proceedings of the
30th International Conference on Neural Information Processing Systems}
(pp. 4338--4346). USA: Curran Associates Inc. Retrieved from
\url{http://dl.acm.org/citation.cfm?id=3157382.3157582}

\leavevmode\hypertarget{ref-Chollet.2015}{}%
Chollet, F. (2015). \emph{Keras}. Retrieved from
\url{https://github.com/fchollet/keras} (Original work published 2015)

\leavevmode\hypertarget{ref-Doya.2000}{}%
Doya, K. (2000). Reinforcement Learning in Continuous Time and Space.
\emph{Neural Computation}, \emph{12}(1), 219--245.
\url{https://doi.org/10.1162/089976600300015961}

\leavevmode\hypertarget{ref-Fechner.etal.1966}{}%
Fechner, G. T., Howes, D. H., \& Boring, E. G. (1966). \emph{Elements of
Psychophysics}. New York: Holt, Rinehart and Winston.

\leavevmode\hypertarget{ref-Graves.etal.2014}{}%
Graves, A., Wayne, G., \& Danihelka, I. (2014). Neural Turing Machines.
Retrieved from \url{http://arxiv.org/abs/1410.5401}

\leavevmode\hypertarget{ref-Graves.etal.2016}{}%
Graves, A., Wayne, G., Reynolds, M., Harley, T., Danihelka, I.,
Grabska-Barwińska, A., \ldots{} Hassabis, D. (2016). Hybrid computing
using a neural network with dynamic external memory. \emph{Nature},
\emph{538}(7626), 471--476. \url{https://doi.org/10.1038/nature20101}

\leavevmode\hypertarget{ref-Grossberg.Merrill.1992}{}%
Grossberg, S., \& Merrill, J. W. (1992). A neural network model of
adaptively timed reinforcement learning and hippocampal dynamics.
\emph{Brain Research. Cognitive Brain Research}, \emph{1}(1), 3--38.

\leavevmode\hypertarget{ref-Grossberg.Schmajuk.1989}{}%
Grossberg, S., \& Schmajuk, N. (1989). Neural Dynamics of Adaptive
Timing and Temporal Discrimination During Associative Learning, 24.

\leavevmode\hypertarget{ref-Hochreiter.Schmidhuber.1997}{}%
Hochreiter, S., \& Schmidhuber, J. (1997). Long Short-Term Memory.
\emph{Neural Computation}, 1735--1780. Retrieved from
\url{http://www.bioinf.jku.at/publications/older/2604.pdf}

\leavevmode\hypertarget{ref-Howard.Shankar.2018}{}%
Howard, M. W., \& Shankar, K. H. (2018). Neural scaling laws for an
uncertain world. \emph{Psychological Review}, \emph{125}(1), 47--58.
\url{https://doi.org/10.1037/rev0000081}

\leavevmode\hypertarget{ref-Howard.etal.2014}{}%
Howard, M. W., MacDonald, C. J., Tiganj, Z., Shankar, K. H., Du, Q.,
Hasselmo, M. E., \& Eichenbaum, H. (2014). A Unified Mathematical
Framework for Coding Time, Space, and Sequences in the Hippocampal
Region. \emph{Journal of Neuroscience}, \emph{34}(13), 4692--4707.
\url{https://doi.org/10.1523/JNEUROSCI.5808-12.2014}

\leavevmode\hypertarget{ref-Howard.etal.2015}{}%
Howard, M. W., Shankar, K. H., Aue, W. R., \& Criss, A. H. (2015). A
distributed representation of internal time. \emph{Psychological
Review}, \emph{122}(1), 24--53. \url{https://doi.org/10.1037/a0037840}

\leavevmode\hypertarget{ref-Jo.2017}{}%
Jo, A. (2017). \emph{Dqn-pytorch: Deep Q Learning via Pytorch}.
Retrieved from \url{https://github.com/AndersonJo/dqn-pytorch} (Original
work published 2017)

\leavevmode\hypertarget{ref-Kaelbling.etal.1998}{}%
Kaelbling, L. P., Littman, M. L., \& Cassandra, A. R. (1998). Planning
and acting in partially observable stochastic domains. \emph{Artificial
Intelligence}, \emph{101}(1), 99--134.
\url{https://doi.org/10.1016/S0004-3702(98)00023-X}

\leavevmode\hypertarget{ref-Kurth-Nelson.Redish.2009}{}%
Kurth-Nelson, Z., \& Redish, A. D. (2009). Temporal-difference
reinforcement learning with distributed representations. \emph{PLoS
One}, \emph{4}(10), e7362.

\leavevmode\hypertarget{ref-Langdon.etal.2018}{}%
Langdon, A. J., Sharpe, M. J., Schoenbaum, G., \& Niv, Y. (2018).
Model-based predictions for dopamine. \emph{Current Opinion in
Neurobiology}, \emph{49}, 1--7.
\url{https://doi.org/10.1016/j.conb.2017.10.006}

\leavevmode\hypertarget{ref-Ludvig.etal.2008}{}%
Ludvig, E. A., Sutton, R. S., \& Kehoe, E. J. (2008). Stimulus
Representation and the Timing of Reward-Prediction Errors in Models of
the Dopamine System. \emph{Neural Computation}, \emph{20}(12),
3034--3054. \url{https://doi.org/10.1162/neco.2008.11-07-654}

\leavevmode\hypertarget{ref-Mnih.etal.2015}{}%
Mnih, V., Kavukcuoglu, K., Silver, D., Rusu, A. A., Veness, J.,
Bellemare, M. G., \ldots{} Hassabis, D. (2015). Human-level control
through deep reinforcement learning. \emph{Nature}, \emph{518}(7540),
529--533. \url{https://doi.org/10.1038/nature14236}

\leavevmode\hypertarget{ref-Paszke.2017}{}%
Paszke, A. (2017). \emph{Pytorch: Tensors and Dynamic neural networks in
Python with strong GPU acceleration}. pytorch. Retrieved from
\url{https://github.com/pytorch/pytorch} (Original work published 2016)

\leavevmode\hypertarget{ref-Sederberg.etal.2008}{}%
Sederberg, P. B., Howard, M. W., \& Kahana, M. J. (2008). A
context-based theory of recency and contiguity in free recall.
\emph{Psychological Review}, \emph{115}(4), 893--912.
\url{https://doi.org/10.1037/a0013396}

\leavevmode\hypertarget{ref-Shankar.Howard.2012}{}%
Shankar, K. H., \& Howard, M. W. (2012). A scale-invariant internal
representation of time. \emph{Neural Computation}, \emph{24}(1),
134--193.

\leavevmode\hypertarget{ref-Shankar.Howard.2013}{}%
Shankar, K. H., \& Howard, M. W. (2013). Optimally fuzzy temporal
memory. \emph{The Journal of Machine Learning Research}, \emph{14}(1),
3785--3812.

\leavevmode\hypertarget{ref-Sozou.1998}{}%
Sozou, P. D. (1998). On hyperbolic discounting and uncertain hazard
rates. \emph{Proceedings of the Royal Society B: Biological Sciences},
\emph{265}(1409), 2015--2020.
\url{https://doi.org/10.1098/rspb.1998.0534}

\leavevmode\hypertarget{ref-Tasfi.2016}{}%
Tasfi, N. (2016). \emph{PyGame-Learning-Environment: PyGame Learning
Environment (PLE) -- Reinforcement Learning Environment in Python}.
Retrieved from
\url{https://github.com/ntasfi/PyGame-Learning-Environment}

\leavevmode\hypertarget{ref-Tiganj.etal.2018}{}%
Tiganj, Z., Cromer, J. A., Roy, J. E., Miller, E. K., \& Howard, M. W.
(2018). Compressed Timeline of Recent Experience in Monkey Lateral
Prefrontal Cortex. \emph{Journal of Cognitive Neuroscience},
\emph{30}(7), 935--950. \url{https://doi.org/10.1162/jocn_a_01273}

\leavevmode\hypertarget{ref-Tiganj.etal.inpress}{}%
Tiganj, Z., Gershman, S. J., Sederberg, P. B., \& Howard, W. (in press).
Estimating scale-invariant future in continuous time. \emph{Neural
Computation}, 24.

\leavevmode\hypertarget{ref-VanEssen.etal.1984}{}%
Van Essen, D. C., Newsome, W. T., \& Maunsell, J. H. R. (1984). The
visual field representation in striate cortex of the macaque monkey:
Asymmetries, anisotropies, and individual variability. \emph{Vision
Research}, \emph{24}(5), 429--448.
\url{https://doi.org/10.1016/0042-6989(84)90041-5}

\end{document}